\documentclass{article}

\usepackage[preprint]{neurips_2026}

\makeatletter
\renewcommand{\@noticestring}{}
\makeatother

\usepackage[utf8]{inputenc}
\usepackage[T1]{fontenc}
\usepackage{amsmath}
\usepackage{amssymb}
\usepackage{graphicx}
\usepackage{booktabs}
\usepackage{array}
\usepackage{microtype}
\usepackage{xcolor}
\usepackage{hyperref}
\usepackage{url}
\usepackage{pgfplots}\pgfplotsset{compat=1.18}
\usepackage{float}
\usepackage{comment} 

\title{ScholarStack: Layered Research Asset Orchestration and Cross-Task Reuse for Scientific Agents}

\author{
  ScholarSeed AI Team\thanks{Full author list available in the Contributions section.}  \\
  DAMO Academy, Alibaba Group \\
  \texttt{wotao.yin@alibaba-inc.com}
}

\begin{document}

\maketitle

\begin{abstract}

Scientific agents support a range of literature-based research tasks, such as retrieval, question answering, evidence-grounded generation, and claim assessment. Most existing systems, however, are organized around individual tasks: the same papers are repeatedly retrieved, segmented and interpreted, and the understanding built in one task is difficult to reuse in the next. We present ScholarStack, a layered research asset framework that compiles a paper collection into reusable, versioned, and provenance-preserving assets at three complementary levels: paper-level statements, domain-level organization, and cross-paper syntheses. A common access interface returns task-specific views at the evidence granularity each task requires, preserving study conditions, source traceability, and verification status. We demonstrate the effectiveness of ScholarStack on four scientific task families spanning ten specific scenarios. 
Quality gains shine on tasks requiring cross-paper evidence, such as multi-paper question answering and literature review generation; query-time token cost also falls as assets are compiled once and reused across tasks. Continuous research chain integration further verifies that task-derived understanding can be efficiently saved, retrieved, and revised. These results suggest that layered research assets can serve as shared infrastructure for scientific agents, shifting literature-based assistance from isolated document processing toward cumulative, evidence-grounded workflows.
\vspace{-0.5em}
\end{abstract}

\section{Introduction}
\label{sec:introduction}
\vspace{-0.4em}
Large language models are increasingly deployed as agents that plan, invoke tools, and execute multi-step workflows. These capabilities are reshaping scientific work: research agents retrieve and synthesize literature, answer technical questions, verify evidence, and assist with proposal and experimental design~\citep{zheng2025automation,ren2025scientific}. Recent systems are also exploring this paradigm toward autonomous experimentation and end-to-end scientific discovery~\citep{lu2026automation,gottweis2026coscientist}. However, scientific research is more than a collection of independent requests or solving isolated tasks. It is a cumulative process: the understandings developed while completing one task should remain interpretable, reusable and revisable for the tasks that follow.

This cumulative requirement is particularly evident when several research activities build on the same literature. A researcher may begin with searching for relevant studies and organizing them into a review, then use that understanding to develop ideas, assess their novelty, and design experiments. Terminology and groups of related studies identified during search can guide the review; relationships and limitations established in the review can guide idea development; and comparisons made during novelty assessment can inform baseline selection and later writing. 
Existing agents support many of these activities~\citep{lala2023paperqa,wang2024autosurvey,li2024chainofideas},
but the intermediate judgments developed in these workflows remain embedded in temporary contexts or task-specific reports, and their evidence and scope are difficult to recover independently. Subsequent tasks must then revisit the same papers and reconstruct reasoning that earlier work has already established.

We refer to this limitation as the \emph{task-specific research bottleneck}: knowledge constructed for one task is difficult to reuse in another without repeated source processing or loss of context. Addressing it requires both \emph{content reuse} and \emph{understanding reuse}. Content reuse retains source statements with their conditions and evidence. Understanding reuse retains the alignments, organizational judgments, and scoped interpretations built from those statements. Reliable reuse further requires explicit boundaries: a result must remain attached to its setting, a shared category must not imply comparability across settings, and a system interpretation must remain distinguishable from a source's assertion. Evidence links and checking status let users inspect a judgment's basis and determine what needs to be re-checked when that basis changes.

Existing approaches provide promising foundations for these reuse requirements. Retrieval-augmented generation and long-context processing improve access to evidence at inference time~\citep{lewis2020rag,gao2023rag}; hierarchical and graph-based retrieval maintain persistent structures such as summary trees and entity graphs~\citep{sarthi2024raptor,edge2024graphrag}; scholarly knowledge bases organize structured research content~\citep{jaradeh2019orkg,raimondi2026scale,weiss2026lacuna,yu2026knows}. Building on these capabilities, reuse across scientific tasks requires each retained object to make explicit what it represents, under which conditions it applies, and what evidence supports it. The requirements differ by the kind of work being reused. Reusing a reported result requires a source statement bound to its study conditions and evidence; reusing the organization of a literature collection requires shared identities, categories, and relations; reusing a cross-paper interpretation requires an account with a declared scope and links to supporting and limiting evidence. These requirements motivate three complementary representations: source-grounded facts, domain organization, and scoped syntheses. Their distinct roles call for separate representations, while their dependencies require explicit links: organizational judgments and syntheses must remain connected to the facts and sources. A common asset model can therefore let tasks reuse each kind of object while retaining the basis needed to interpret and check it.


We introduce \textbf{ScholarStack}, a layered research asset framework for scientific agents. It \emph{compiles} heterogeneous research materials into three linked layers: L1 records source-grounded scientific facts together with their study context and evidence references; L2 organizes them through a domain taxonomy, canonical entities, category assignments, and typed relations; L3 forms scoped cross-paper syntheses that state their coverage and evidence basis explicitly. 
A common access interface supports retrieving, selecting, expanding, comparing, and synthesizing knowledge, and returns typed knowledge views over these layers. This allows heterogeneous downstream tasks, from literature retrieval and question answering to evidence-grounded generation and claim assessment, to reuse the same research assets while selecting the evidence granularity they need. The resulting assets can remain revisable as materials change. Selected task-derived objects may optionally be registered after identity and reference checks, with semantic verification recorded separately.

We instantiate and evaluate the framework on four task families spanning a literature-based research workflow: literature retrieval, scientific question answering, evidence-grounded generation, and claim and consistency assessment. These tasks draw on the layers differently. For example, single-paper question answering exercises primarily the source-grounded facts of L1, whereas literature review generation benefits further from L2 organization and L3 syntheses. Assets compiled once from the source papers serve all tasks through the shared interface. We observe general performance and token efficiency benefits of ScholarStack across ten specific research scenarios, while the largest gains appear on tasks that require cross-paper evidence, such as multi-paper question answering and literature review generation. Continuous research chains further demonstrate understanding reuse and revision at lower per-chain token cost. Our contributions are as follows:
\begin{itemize}
    \setlength{\itemsep}{-1pt}
    \item We characterize the \emph{task-specific research bottleneck}: repeated source processing and reconstruction of prior judgments across tasks. We distinguish reuse of source-grounded content from derived understanding, and identify the evidence and scope needed for both.
    \item We derive a layered research asset model from these requirements: source-grounded facts, domain organization, and scoped syntheses remain distinct but linked, with a common access interface and registration mechanisms for versioned, reusable objects.
    \item We evaluate ScholarStack across ten task settings in four families and continuous research chains, demonstrating quality and efficiency benefits alongside traceable reuse and revision of task-derived understanding.
\end{itemize}


\section{Related Work}
\label{sec:related}

\paragraph{Scientific agents for literature-based research.}
Large language models are increasingly deployed as \emph{scientific agents} that automate stages of the research lifecycle, and recent surveys chart their rapid progress across literature synthesis, question answering, evidence verification, and experimental design~\citep{zheng2025automation,ren2025scientific}. Beyond literature-based assistance, agents have begun to drive autonomous experimentation and end-to-end scientific discovery~\citep{lu2026automation,gottweis2026coscientist}. Closest to our setting are agents built specifically for literature-based research. PaperQA retrieves scientific literature and assesses evidence relevance to answer research questions~\citep{lala2023paperqa}, OpenScholar synthesizes the literature through retrieval-augmented generation with citations~\citep{asai2026openscholar}, and long-form systems such as STORM organize retrieved material into structured, grounded reports~\citep{shao2024storm}. These systems are highly capable at individual tasks, but each is organized around a single request. The understanding built while completing one task, including aligned terminology, cross-paper relationships, and verified findings, remains embedded in a transient context or a task-specific output and is rarely registered in a persistent, reusable form. ScholarStack instead compiles this intermediate understanding into versioned, provenance-preserving assets, so that knowledge constructed for one task becomes infrastructure for the next rather than being reconstructed from the sources at each step.

\vspace{-0.5em}
\paragraph{Retrieval augmentation and hierarchical indexing.}
Retrieval-augmented generation (RAG) grounds model outputs in external corpora at inference time and has become a standard paradigm for knowledge-intensive tasks~\citep{lewis2020rag,gao2023rag}. To move beyond flat passage retrieval, hierarchical and graph-based methods build persistent index structures over a corpus. RAPTOR recursively summarizes text into a tree that supports retrieval at multiple levels of abstraction~\citep{sarthi2024raptor}, and GraphRAG links an entity graph with community summaries to answer corpus-level questions~\citep{edge2024graphrag}. A related line reasons directly over knowledge-graph structure, as in Think-on-Graph's traversal of relation paths~\citep{sun2024tog} and G-Retriever's subgraph retrieval for textual-graph question answering~\citep{he2024gretriever}. These approaches demonstrate the value of access across levels of abstraction, but their primary abstraction is a retrieval structure optimized to fetch relevant context, and it does not separate the distinct evidential roles that reusable research knowledge must carry, namely what an individual source states and under which conditions, how research objects relate within a domain, and what several studies jointly support. ScholarStack keeps these three roles distinct yet linked and exposes them through typed operations, rather than folding them into a single summary tree or entity graph that is queried anew for each request.

\vspace{-0.5em}
\paragraph{Scholarly knowledge bases and structured research representations.}
A complementary line represents scholarly knowledge in structured, machine-actionable form. The Open Research Knowledge Graph (ORKG) curates paper contributions as structured, comparable descriptions to support their reuse~\citep{jaradeh2019orkg}. SCALE organizes scientific terminology into reusable concepts within a disciplinary taxonomy~\citep{raimondi2026scale}, Lacuna links paper-level content to research directions and open problems~\citep{weiss2026lacuna}, and Knows exposes agent-native structured representations of research content~\citep{yu2026knows}. More broadly, coupling language models with knowledge graphs has been framed as a spectrum from KG-enhanced models to synergized systems in which the two are mutually reinforcing~\citep{pan2024unifying}. Each of these commits to a single primary representation, whether a contribution graph, a concept hierarchy, a research map, or a typed entry, and while several already record provenance and support reuse, none jointly maintains source-grounded facts, domain organization, and scoped cross-paper syntheses as separate-but-linked assets under shared provenance and revision semantics. ScholarStack targets this integration problem by maintaining a single evolving asset collection that preserves the three evidential roles, tracks their provenance and versions, and serves task-specific views over the same knowledge, so that a sequence of related research activities accumulates rather than repeats.

\section{Method: Layered Research Asset Orchestration and Reuse}

\label{sec:method}

ScholarStack separates the construction of reusable research assets from their use in downstream tasks. We first derive the asset design from research requirements, then describe its knowledge representations and construction mechanisms. Finally, we explain how agents access these assets for individual tasks and preserve task-derived understanding for subsequent research.

\subsection{Research Requirements and Asset Design}

\label{sec:framework}

Research tasks repeatedly revisit the same literature, but the knowledge worth preserving extends beyond retrieved passages. Answering a question requires recovering what is reported under which conditions; comparing methods requires identifying related research and checking their assumptions, measurements, and evaluation settings; synthesis requires clarifying what multiple studies jointly support and within what scope. When such judgments remain embedded in task-specific outputs, subsequent tasks need to reconstruct their conditions, organization, and evidential basis.

These requirements imply that reusable research assets must preserve three complementary forms of information (Table~\ref{tab:research_requirements}). Source-grounded statements preserve what an individual paper asserts together with its conditions and source locations (L1); domain organization preserves identities, categories, and relations together with the basis and uncertainty of these judgments (L2); scoped syntheses preserve cross-paper interpretations together with their coverage and supporting, opposing, or limiting evidence (L3). The distinction reflects different evidential responsibilities: fidelity to a source, justification of an organizational judgment, and support for a scoped interpretation. Shared identifiers and evidence references connect these objects without treating them as interchangeable forms of knowledge.

\begin{table}[htbp]
\centering
\small
\caption{Research requirements and corresponding reusable assets.}
\label{tab:research_requirements}
\renewcommand{\arraystretch}{1.08}
\setlength{\tabcolsep}{5pt}

\begin{tabular}{
@{}
>{\raggedright\arraybackslash}p{0.19\linewidth}
>{\raggedright\arraybackslash}p{0.43\linewidth}
>{\raggedright\arraybackslash}p{0.32\linewidth}
@{}
}
\toprule
Research need
& Reusable assets
& Reuse benefits \\
\midrule

Grounded reading and verification
&
\textbf{L1:} Source-grounded facts with study conditions, source locations, and checking states.
&
Reduce repeated source processing while preserving the context of reported findings.
\\
\addlinespace[3pt]

Organization and comparison
&
\textbf{L1 + L2:} Canonical entities, categories, and relations linked to result-specific assumptions, measurements, and settings.
&
Reuse organizational judgments while avoiding comparisons across incompatible settings.
\\
\addlinespace[3pt]

Cross-paper synthesis
&
\textbf{L3 (grounded in L1 + L2):} Scoped syntheses built over organized research objects and linked to source-grounded evidence.
&
Reuse scoped interpretations without extending conclusions beyond the examined evidence.
\\
\addlinespace[3pt]

Inspection and revision
&
\textbf{Across layers:} Evidence dependencies, identifiers, versions, and checking states.
&
Preserve justification and identify what requires rechecking when evidence changes.
\\

\bottomrule
\end{tabular}
\end{table}

\begin{figure}[t]
    \centering
    \includegraphics[width=\linewidth]{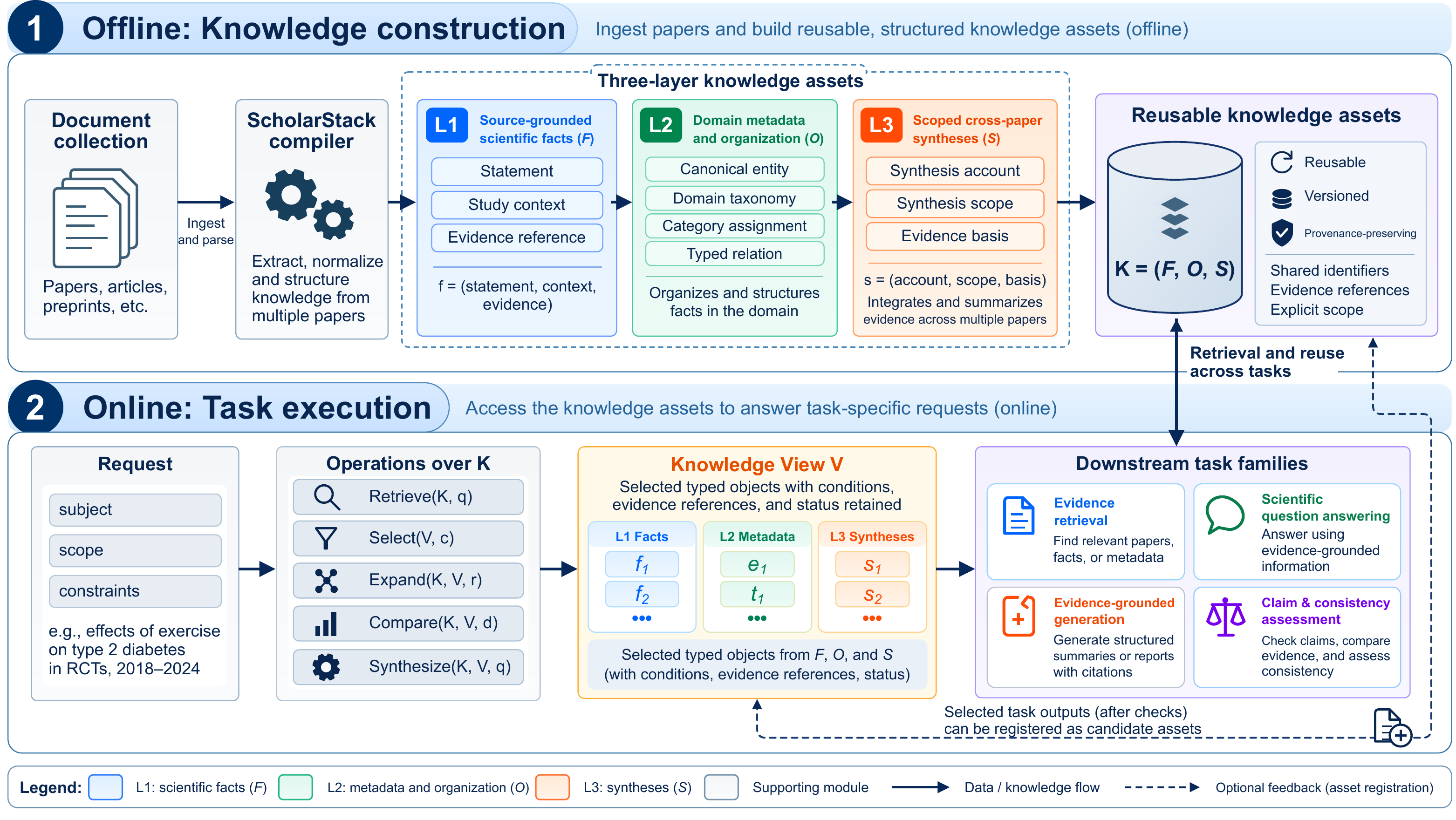}
    \vspace{-0.5em}
    \caption{
        Overview of ScholarStack.
        Knowledge construction compiles documents into
        source-grounded facts (L1), domain organization (L2), and
        scoped cross-paper syntheses (L3), linked by shared identifiers
        and evidence references.
        During task execution, agents select and combine the required
        objects into a Knowledge View, retaining their conditions,
        evidence references, and recorded verification states.
        Different tasks use different subsets of the layers and
        operations.
        The feedback path denotes optional registration of task-derived
        candidates after the applicable checks; registration does not
        itself establish semantic verification.
    }
    \vspace{-0.5em}
    \label{fig:main_method}
\end{figure}

ScholarStack realizes this design as three linked knowledge layers (Fig.~\ref{fig:main_method}): L1 captures source-grounded scientific facts, L2 represents domain organization, and L3 represents scoped cross-paper syntheses. The layers are complementary rather than a mandatory processing sequence for every task. For example, L2 may identify two methods as belonging to the same category, but whether their empirical results are directly comparable must still be determined from the conditions retained in L1; an L3 synthesis may integrate those results only within a scope justified by its evidence.

During task execution, the agent selects and combines the required objects into a task-specific Knowledge View. Different tasks can therefore select the layers and evidence granularity they require: a single-paper question may rely mainly on L1, a method comparison may combine L2 with L1 conditions, and a literature review may additionally draw on L3 syntheses. The following sections formalize these representations and describe how they are constructed and accessed.

\subsection{Knowledge Representation and Construction}
\label{sec:representation}

The knowledge layer compiles a document collection into a layered representation
\begin{equation}
K=(F,O,S),
\end{equation}
where $F$ is a collection of source-grounded Scientific Facts (L1), $O$ is the domain organization that aligns and relates them (L2), and $S$ is a collection of Scoped Syntheses that interpret them jointly (L3). These representations carry different evidential commitments: facts preserve what a source states and under which conditions; domain organization establishes identities, categories, and relations with an explicit basis; syntheses integrate findings within a declared scope. Keeping these roles distinct allows each to be checked and revised according to the kind of knowledge it represents.

\vspace{-0.5em}
Attributes and taxonomy dimensions vary with the discipline and study type, while these three roles remain distinct. Appendix~\ref{app:cross-layer-example} traces one research topic from source sentences through L1 facts and L2 organization to an L3 synthesis.

\subsubsection{L1: Source-Grounded Scientific Facts}
\label{sec:l1}

\paragraph{Representation.}
L1 directly anchors scientific statements to source sentences and provides the evidential basis for higher-layer objects. A \textbf{Scientific Fact} $f\in F$ records one statement from one source together with what is needed to interpret it, $f=(\mathrm{statement},\mathrm{context},\mathrm{evidence})$: the statement identifies a subject and what the source reports about it, the context records the assumptions and conditions under which the report holds, and the evidence field locates the supporting sentence or sentences. Statements carry a type, such as a finding, a method description, a hypothesis, or a limitation, and a single paper typically contributes facts of several kinds. A quantitative fact keeps the reported value bound to the subject and the setting it was measured in, and every fact carries a verification status recording whether it has been checked against its cited sentences. In every case a fact records what the source asserts rather than certifying its truth.

\vspace{-0.6em}
\paragraph{Construction.}
Facts are extracted by a language model guided by schemas for the study type, so that quantitative, qualitative, and theoretical content keep their corresponding conditions. Structural checks validate the required fields and source references. Semantic verification assesses whether a statement is attributed to the correct subject and preserves the content and qualifications of its cited sentences. Failed and unresolved semantic checks remain explicit in the fact's verification status.

\vspace{-0.6em}
\paragraph{Role in access.}
Recorded contexts let tasks distinguish relevant facts that concern different populations, assumptions, or evaluation settings. L1 supplies both reusable content and the conditions governing its comparison. Appendix~\ref{app:l1} presents a detailed example for a real paper.

\subsubsection{L2: Domain Metadata and Organization}
\label{sec:l2}

\paragraph{Representation.}
Facts from different papers do not line up by themselves: the same object may appear under different names, and $F$ alone does not specify their categories or relationships. The organization $O=(T,E,A,R)$ supplies this structure. A \textbf{Domain Taxonomy} $T$ defines categories and membership criteria along field-specific dimensions, such as method family, problem setting, or guarantee type. \textbf{Canonical Entities} $E$ give a shared identity to mentions identified as the same method, dataset, metric, or other research object. \textbf{Category Assignments} $A$ place an entity in a category of $T$, cite supporting facts, and record whether membership is author-declared or inferred and with what certainty. Assignments derived from a particular result retain that result's conditions; labels from different results cannot be combined into a single guarantee. \textbf{Typed Relations} $R$ connect identified objects through named relationships, such as evaluation on a dataset, comparison against a baseline, or improvement over an earlier method, with payloads such as reported results or gains.

\paragraph{Construction.}
The three derived kinds are built from the extracted facts rather than from raw text: mentions are merged by deterministic rules with model adjudication for ambiguous cases, assignments must cite supporting facts, and relations are anchored in the extracted records that connect their endpoints. The taxonomy may be supplied by experts or prepared from the collection. Author-declared and inferred membership stay distinguishable, and unresolved cases remain uncertain rather than forced into a category. Construction also runs top-down: the taxonomy guides what extraction attends to and along which dimensions findings are compared, and facts that resist classification prompt taxonomy revision or re-extraction.

\paragraph{Role in access.}
$O$ supports selection by category, traversal along relations, and comparison along the dimensions of $T$. Membership in a common category establishes relevance, while comparability is still decided by the contexts recorded in $F$: two methods in the same family need not share assumptions, and shared membership alone does not justify aggregating their results. Appendix~\ref{app:l2} shows how the same paper's facts support domain assignments and relations.

\subsubsection{L3: Scoped Cross-Paper Synthesis}
\label{sec:l3}

\paragraph{Representation.}
A \textbf{Scoped Synthesis} $s\in S$ records an interpretation formed across papers, $s=(\mathrm{account},\mathrm{scope},\mathrm{basis})$. It may describe a method family, a qualified disagreement, or an open problem within the covered studies. The account states the interpretation; the scope declares the topic, applicable conditions, and studies considered; the basis cites registered facts and, where relevant, category assignments. Each cited object is marked as supporting, opposing, or limiting, exposing both the grounds for the interpretation and its boundaries.

\paragraph{Construction.}
Synthesis uses $O$ to organize and select related material, while its account is grounded in direct citations to the selected facts. Conflicting findings about the same question under comparable conditions are recorded as a disagreement. Differences attributable to study context are described separately, and unresolved uncertainty remains explicit. The generating model may cite only registered facts and assignments; candidates whose citations cannot be resolved are rejected before registration. Reference validation establishes that the basis is identifiable; semantic review separately assesses whether it supports the account and its declared scope. The scope describes the evidence considered, not complete coverage of a field.

\paragraph{Role in access.}
Tasks can retrieve an existing synthesis with its basis or request a new account from selected evidence. A new account remains a candidate until registration. Appendix~\ref{app:l3} illustrates a method-family synthesis based on saved records, including its scope, supporting and limiting evidence, and review state.

\subsubsection{Cross-Layer Integration and Evidence Traceability}
\label{sec:integration}

Shared identifiers link assignments to supporting facts, relations to the extracted records that connect their endpoints, and syntheses to facts and relevant assignments. These references connect derived objects to source sentences and make their evidential dependencies inspectable.

Registration turns a knowledge object into a persistent \textbf{Research Asset} by validating required fields, identity, and reference integrity, reusing existing matches, and versioning new or revised objects. Structurally invalid outputs remain outside $K$. Semantic verification is recorded separately: failed or unresolved objects remain inspectable, and new syntheses do not inherit verification from their evidence. If a source or asset is revised, its dependents require rechecking; earlier versions preserve the basis of prior outputs. Task-derived knowledge objects may also be registered through the same mechanism for reuse in subsequent tasks. The construction cost and its near-linear scaling with corpus size are analyzed in Appendix~\ref{app:continuous-cost}.

\subsection{Task-Oriented Knowledge Access and Application}
\label{sec:application}

A task accesses $K$ through a \textbf{Knowledge View} $V$: a selection of typed objects retaining their conditions, evidence references, and verification states. Table~\ref{tab:knowledge-operations} defines five operations for constructing and using such views. The agent chooses operations and layers according to the request's subject, scope, and constraints; there is no fixed sequence or requirement to use all three layers. A paper-scoped task may also receive the designated paper's complete fact collection directly.

Task outputs retain the evidential distinctions established in Section~\ref{sec:representation}. Comparison exposes differences in conditions without implying a ranking, and newly generated interpretations remain distinct from registered syntheses. Evidence references locate source sentences; when original-text access is allowed, the agent may read surrounding passages to check context. Knowledge-only settings rely on the recorded evidence and disclose its limitations. Missing evidence indicates limited coverage of the available view; it does not establish that no relevant work exists.

\begin{table}[htbp]
\centering
\small
\caption{Core operations over the knowledge layer. $K=(F,O,S)$ comprises source-grounded facts, domain organization, and scoped syntheses; $q$ is a retrieval or synthesis request; $V$ is a selected view of typed knowledge objects, and $V^{\prime}$ is the view returned by selection or expansion; $c$ specifies selection constraints, $r$ the relations or evidence links to follow, and $d$ comparison dimensions; $C$ is a Comparison Result, and $s$ a candidate Scoped Synthesis. The signatures specify semantic inputs and outputs rather than a fixed execution policy.}
\label{tab:knowledge-operations}
\begin{tabular}{p{0.27\linewidth} p{0.67\linewidth}}
\toprule
Operation & Output and semantics \\
\midrule
$\mathbf{Retrieve}(K,q)\rightarrow V$ & Locate recorded facts, domain objects, or existing syntheses relevant to the request, retaining their sources and status. \\
$\mathbf{Select}(V,c)\rightarrow V'$ & Filter the view by explicit conditions, distinguishing confirmed matches from mismatches and missing or inapplicable information. \\
$\mathbf{Expand}(K,V,r)\rightarrow V'$ & Supplement the view with objects connected through the specified relations or evidence links, retaining the connections to the selected objects. \\
$\mathbf{Compare}(K,V,d)\rightarrow C$ & Organize findings by the requested comparison dimensions, exposing shared and differing conditions and identifying unresolved comparability. \\
$\mathbf{Synthesize}(K,V,q)\rightarrow s$ & Generate a scoped interpretation from the selected evidence, recording its supporting, opposing, and limiting basis as a candidate synthesis. \\
\bottomrule
\end{tabular}
\end{table}

\subsubsection{Literature Retrieval}
\label{sec:retrieval}

\begin{figure*}[t]
\centering
\includegraphics[width=0.98\textwidth]{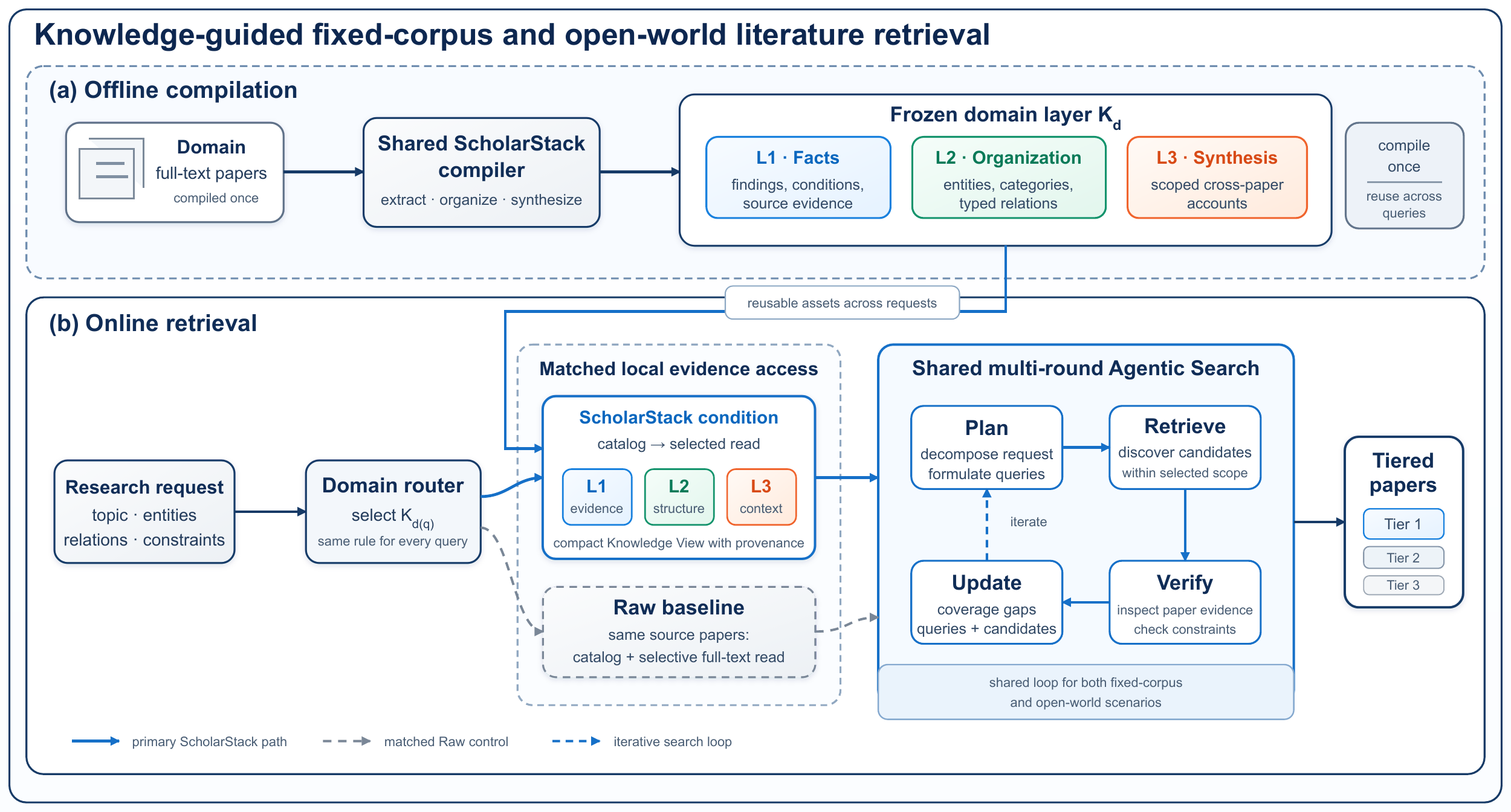}
\caption{Knowledge-guided agentic literature retrieval. A compiled Knowledge View informs search planning and evidence selection. Candidate papers are checked against the request using retrieved source evidence, which also supports the final ranking. Fixed-corpus and open-world settings differ in the permitted search scope.}
\label{fig:open-world-retrieval}
\end{figure*}

For literature search, the view guides query formulation and evidence selection: L1 supplies findings and conditions, while relevant L2 organization and L3 syntheses suggest terminology and related approaches. The agent discovers candidate papers, checks them against source evidence, and refines its search as needed (Fig.~\ref{fig:open-world-retrieval}). Final ranking relies on the retrieved papers.

\paragraph{Fixed-corpus retrieval.}
Given a request and a fixed collection, $\mathbf{Retrieve}$ and $\mathbf{Select}$ assemble relevant knowledge. Source links and knowledge-guided queries identify papers within that collection. The agent checks abstracts and relevant passages against the requested conditions and returns ranked candidates, distinguishing supported matches from unresolved details.

\paragraph{Open-world retrieval.}
Given a request, the agent uses a view from an existing domain collection to formulate searches over external scholarly sources. It consolidates candidate identities, verifies the requested conditions, and returns papers with their supporting evidence. The compiled collection remains fixed during this search; newly discovered papers need not already belong to it.

\paragraph{Novelty assessment.}
\label{sec:m-novelty}
Given a proposed contribution and prior-work corpus, the agent decomposes the contribution by problem, mechanism, and setting. Relevant L1 facts supply comparison evidence, and L2 Canonical Entities align naming variants. $\mathbf{Compare}$ identifies overlaps and residual differences with cited support. The output is bounded by the corpus and evidence examined; failure to find an overlap does not establish global novelty.

\subsubsection{Scientific Question Answering}
\label{sec:qa}

\begin{figure}
\centering
\includegraphics[width=1\linewidth]{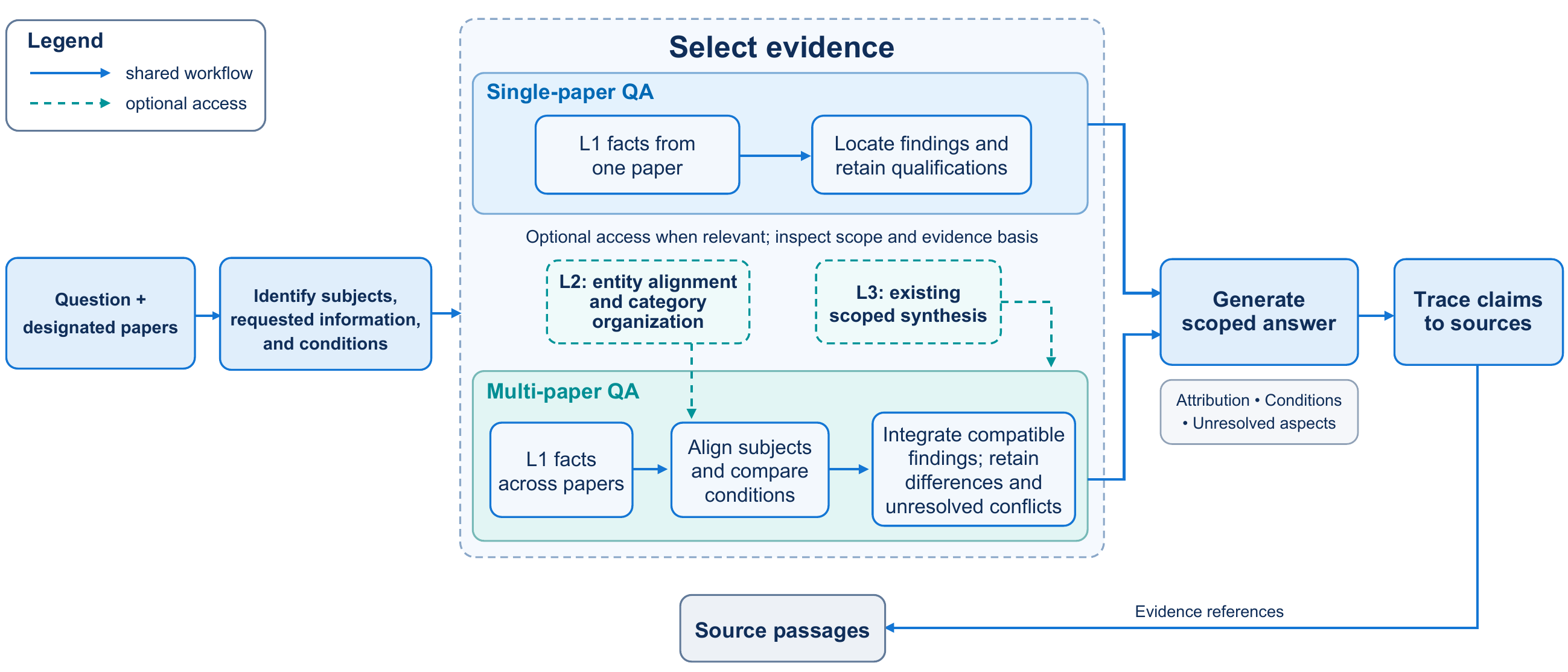}
\caption{Scientific question answering over designated papers. Single-paper QA selects source-grounded facts; multi-paper QA additionally aligns subjects and compares study conditions. L2 organization and existing L3 syntheses contribute when relevant. Evidence references trace answer claims through supporting facts to source sentences; access to surrounding source text depends on the task setting.}
\label{fig:scientific-qa}
\end{figure}

Given a question and designated papers, the agent identifies the requested information and conditions, assembles evidence, and generates an answer with citations (Fig.~\ref{fig:scientific-qa}). Source-reported statements remain distinguishable from interpretations introduced during answering.

\paragraph{Single-paper question answering.}
Source-grounded facts and the sentence-level source content indexed with them supply the answer evidence from the designated paper, located at a finer granularity than whole passages. When available, L2 Canonical Entities assist name alignment without expanding the evidence scope to other papers; L3 is generally unnecessary. The answer preserves the facts' contexts and evidence references. If the selected evidence is insufficient, the agent checks the source when permitted or reports the unresolved information within the available view.

\paragraph{Multi-paper question answering.}
\label{sec:multi-paper-qa}
Facts retain their paper identities and contexts. L2 entities align subjects, while $\mathbf{Compare}$ organizes findings by question-specific dimensions, such as assumptions, settings, and outcomes. The answer combines compatible findings and preserves contextual differences and unresolved conflicts. An existing L3 synthesis may contribute if its scope matches the question; $\mathbf{Expand}$ exposes its basis. Otherwise, $\mathbf{Synthesize}$ can form a candidate account from the selected facts. Answer-time interpretations do not automatically become registered L3 assets.

\subsubsection{Evidence-Grounded Generation}
\label{sec:generation}

\begin{figure}[t]
    \centering
    \includegraphics[width=\linewidth]{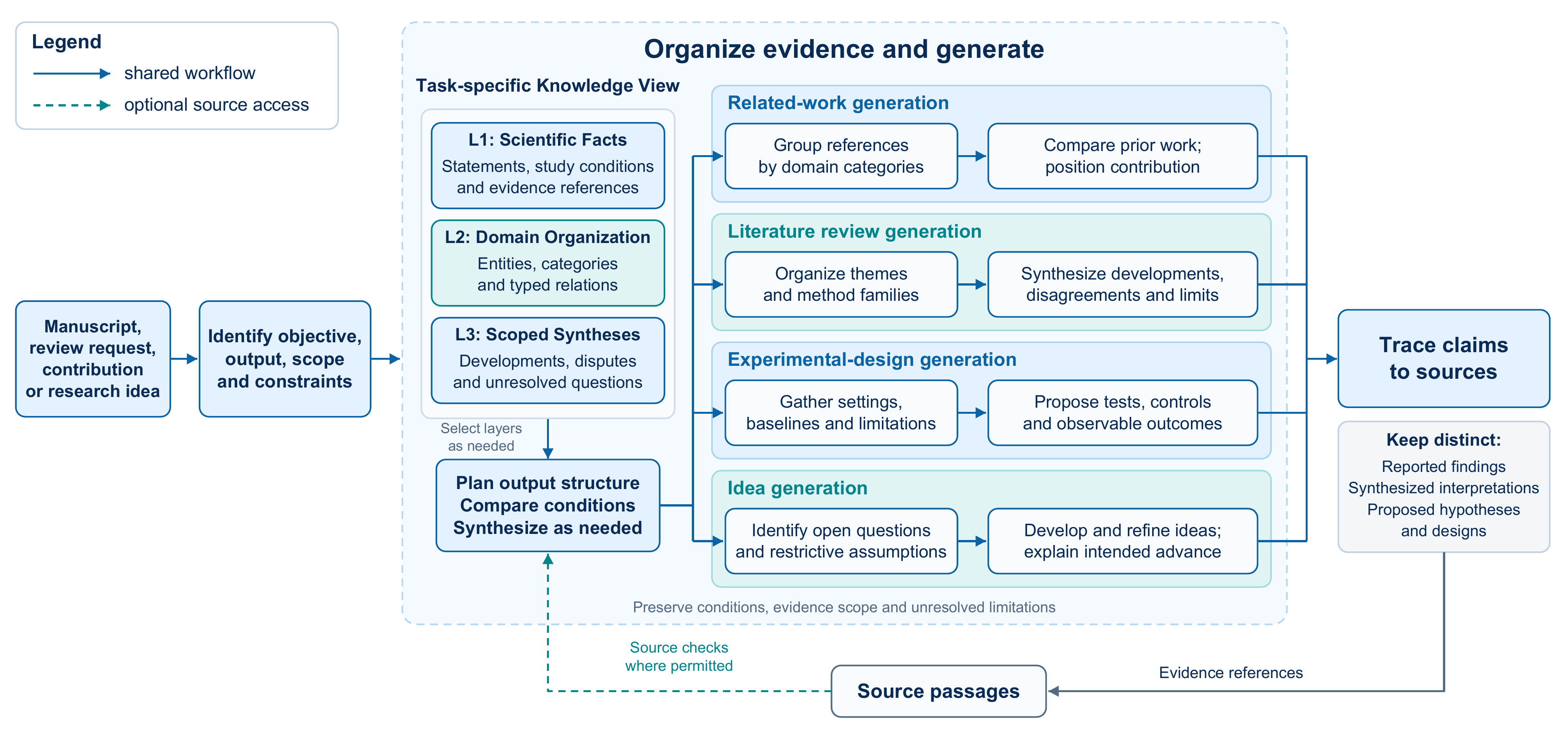}
    \caption{Evidence-grounded generation. A task-specific Knowledge View supports planning, comparison, and synthesis for related-work, review, experimental-design, and idea generation. Outputs retain study conditions and evidence references and distinguish literature-based interpretations from proposed hypotheses and design choices.}
    \label{fig:evidence-grounded-generation}
\end{figure}

Given a writing or research objective, the agent organizes a view into themes, comparisons, or questions (Fig.~\ref{fig:evidence-grounded-generation}). L2 categories and relations guide organization; L1 facts substantiate claims; L3 syntheses provide scoped interpretations. $\mathbf{Compare}$ aligns conditions, and $\mathbf{Synthesize}$ supplies an additional account when needed. Outputs distinguish literature-supported statements and interpretations from proposed hypotheses and design choices, which do not inherit verification from their motivating evidence.

\paragraph{Related-work generation.}
Given a focal manuscript and reference collection, the agent identifies the target's research problem, approach, and claimed contribution. L2 Category Assignments group references into research threads. Typed Relations capture support, contrast, and extension among studies. L1 facts provide evidence for comparing methods, assumptions, and findings while preserving their study conditions. L3 Scoped Syntheses provide cross-paper context within an explicit scope. The agent applies $\mathbf{Compare}$ to organize the evidence along dimensions relevant to the target. It then writes a coherent account that relates prior work to the target's approach and claimed contribution. Substantive claims about prior work cite the supplied references.


\paragraph{Literature review generation.}
\label{sec:review-method}
For a survey request, ScholarStack first retrieves a task-specific literature collection and compiles its knowledge assets, then constructs a view for writing. Existing assets can be reused; updates are needed only when the intended coverage requires additional or revised material. The writer uses the L2 Domain Taxonomy and Category Assignments to organize the field, L3 Scoped Syntheses to characterize developments and disagreements, and L1 facts with their source sentences to substantiate technical claims. It links the outline and central assertions to the selected facts and syntheses through their evidence references, then checks attribution, quantities, comparison scope, and citation placement. This top-down organization followed by bottom-up source checking exposes coverage gaps and allows provisional classifications and interpretations to be qualified or rejected. 

\paragraph{Experimental-design generation.}
\label{sec:m-expdesign}
Given a contribution's claims, the agent gathers L1 facts about relevant procedures, baselines, observables, and limitations, using L2 relations and L3 accounts to organize them. For each claim, the plan specifies a change, comparator, observations that would support or weaken it, and applicable conditions. Reported settings provide the rationale; new design choices are marked as proposals. 

\paragraph{Idea generation.}
\label{sec:m-idea-generation}
Given a topic or preliminary idea, the idea-generation skill identifies the research problem and resource constraints. It develops candidate directions and assesses their scientific value and feasibility. Through internal iteration, it strengthens each candidate's rationale and outlines how the proposed contribution could be tested. In ScholarStack, the agent applies this skill using evidence accessed through the asset layer: L1 facts and study contexts, L2 relations, and relevant L3 syntheses. These assets support comparisons with prior approaches and help assess whether candidate ideas fit the conditions of the supporting evidence. The final ideas state their proposed contributions, rationale, and supporting references.

\subsubsection{Claim and Consistency Assessment}
\label{sec:assessment}

This task family assesses assertions against available evidence. Here we instantiate it as claim verification against a designated cited paper, separately from the extraction verification used during knowledge construction.

\paragraph{Scientific claim verification.}
\label{sec:m-verifier-claim}
Given a claim and paper, the verifier receives the paper's complete L1 fact collection, without query-time retrieval or L2/L3 processing. Statements and contexts support a relation label and rationale distinguishing direct support, overstatement, topical similarity without support, and irrelevance~\citep{wadden2020scifact}. Missing qualifications can affect the judgment, and incomplete evidence references limit traceability~\citep{wadden2022multivers}.

\subsubsection{Continuous Research}
\label{sec:m-continuous-research}

Given a research objective spanning multiple tasks, the agent assembles a Knowledge View for each task from the shared asset collection, including relevant objects retained from earlier tasks. $\mathbf{Retrieve}$ and $\mathbf{Expand}$ expose prior findings and judgments with their conditions and evidence, supporting reuse in subsequent work. Task order and layer selection depend on the research objective.

Tasks may also contribute new or revised assets through optional registration. For example, literature review generation may identify supplementary L1 facts, L2 entity aliases, or L3 scoped syntheses; counterevidence encountered during novelty assessment may motivate revising an earlier synthesis. These contributions depend on the evidence encountered, with no fixed task-to-layer mapping. Later tasks can access the registered objects and their revision states. Section~\ref{sec:continuous-research} evaluates one five-stage instantiation and dependent follow-up chains.

\section{Experiments}
We evaluate ScholarStack across diverse scientific tasks and continuous research chains, focusing on task quality, token efficiency, and the reuse of prior understanding. We also examine how different knowledge layers contribute to these outcomes.

\subsection{Experimental Setup}
We evaluate ScholarStack across four task families: literature retrieval, scientific question answering, evidence-grounded generation, and claim and consistency assessment. Unless otherwise specified, we use Qwen3.8-Max as the base model and Codex as the agent framework. Within each matched comparison, conditions share the source collection, base model, task instructions, and execution budget, subject to the task-specific access and prompting variants. Throughout the experiments, \textbf{Full-text} denotes the baseline that accesses the original paper texts, and \textbf{ScholarStack} denotes our method, which accesses research assets compiled from the same sources. 

Evaluation uses task-specific quality metrics and, where available, query-time resource usage. Unless noted otherwise, token counts include model input and output. Construction costs, reuse assumptions, and further implementation details are reported in the appendices.

\subsubsection{Literature Retrieval}

\paragraph{Fixed-corpus retrieval.}
We evaluate fixed-corpus literature retrieval on 50 queries sampled from LitSearch~\citep{ajith2024litsearch}. ScholarStack uses assets compiled from 900 representative source papers, while Full-text accesses the same sources as raw text. Both follow a four-stage protocol comprising planning, expansion, verification, and ranking under matched search and reading budgets. We additionally compare three deterministic baselines: BM25, Dense embeddings, and Hybrid retrieval. We report macro-averaged gold-paper recall at ranks 1, 5, and 10, MRR@40, and mean time and tokens per query. Corpus construction, baseline implementations, execution limits, and metric definitions are detailed in Appendix~\ref{app:fixed-corpus-retrieval}.

\paragraph{Open-world retrieval.}
In this task, we evaluate whether ScholarStack improves the high-priority results of an agent that can search beyond the layer's source collection. We use the open-ended portion of SAGE~\citep{hu2026sage}, which associates each research-style query with two \emph{most relevant} seed papers and a set of \emph{relevant} shared references. We select 89 queries covering computer science, natural science,  and healthcare domains. Each query has two \emph{most relevant} seed papers and additional \emph{relevant} references. A query-independent collection of approximately 1,750 full-text papers per domain supplies the local context; benchmark domain annotations route queries to the corresponding collection. Full-text receives a title-and-abstract catalog followed by selected full-text passages, whereas ScholarStack receives a structured catalog followed by selected L1/L2/L3 records. Both use one catalog call and one read call within a 12,000-character budget, followed by the same multi-round Agentic Search workflow over the public Web and an internal scholarly index. We score the explicit Tier~1 recommendations, averaging 4.95 papers per query, using macro recall, precision, and F1, together with macro and micro weighted recall (weights of 2 and 1 for the two relevance levels). Further details are provided in Appendix~\ref{app:open-corpus-retrieval}.


\paragraph{Novelty assessment.}
\label{sec:exp-novelty}

We evaluate whether a system can distinguish substantive overlap with prior work from topical similarity.
For this task, we build a cross-domain benchmark from a citation-linked scientific corpus spanning 12 fields. It contains 40 overlap-present targets and 135 gold overlaps. Ground truth consists of a target's prior work identified through citation relations; same-domain non-cited papers serve as hard distractors. Both methods share the base model, tool budget, and closed-book setting, differing only in knowledge access: Full-text applies our novelty assessment method while reading the original related-paper texts, and ScholarStack applies the same method over assets compiled offline from those texts. Web search is disabled. We report weighted identification F1, using an LLM matcher and weights of $1.0$ for core overlap axes and $0.3$ for peripheral axes, together with tokens per query. Appendix~\ref{app:novelty} provides matcher, cost-accounting, ablation, and weighting-sensitivity details.

\subsubsection{Scientific Question Answering}
\label{sec:exp-single-qa-setup}

\paragraph{Single-paper QA.}
We evaluate answering from one designated paper on QASPER~\citep{dasigi-etal-2021-dataset} (416 papers and 1{,}428 questions) and PeerQA~\citep{baumgartner-etal-2025-peerqa} (208 papers and 579 questions). This setting primarily tests access to local, source-grounded evidence. Full-text receives the entire paper; RAG retrieves fixed passages; ScholarStack retrieves structured facts and sentence-level evidence. Conditions share the answering prompt, and RAG and ScholarStack use the same semantic retriever. For QASPER, we report official Answer-F1, its answer-type breakdown, and Evidence-F1. For PeerQA, we report answerability metrics over 495 questions and generation Rouge-L against free-form answers (FF) and annotated evidence (AE) over 245 questions. Mean input tokens per question measure query-time cost. Appendix~\ref{app:single-qa} provides benchmark composition and evaluation details and additional analyses.

\paragraph{Multi-paper QA.}

We evaluate our method on MDAQA~\citep{huang-etal-2025-MDAQA}, which contains 797 questions, each referring to at least two articles. Each question is accompanied by gold answers, from which we identify the corresponding gold papers. Using the same ordinary prompt without additional retrieval, we compare Full-text, which provides the complete parsed texts of the gold papers, with ScholarStack. GPT-6 Astra with high reasoning effort evaluates all candidate answers anonymously against a fixed evidence package, with randomized candidate identifiers and presentation order. Condition denotes the input setting used to generate each answer and is hidden from the evaluator. Correctness (Corr.) assesses evidential support and accurate attribution of subjects, values, and conditions. Completeness (Comp.) measures coverage of the question, while cross-paper synthesis (Synth.) assesses evidence-based integration and comparison across papers. Calibration (Cal.) measures whether the strength of assertions and stated limitations are consistent with the evidence. Directness (Dir.) assesses relevance and the avoidance of redundancy. Tokens denote the mean number of input and output tokens per question, and Sec. denotes the mean generation time in seconds per question. Costs include answer generation only, excluding knowledge construction and judging. Additional details are provided in Appendix~\ref{app:multi-paper-qa}.
\subsubsection{Evidence-Grounded Generation}
\label{sec:setup-generation}

\paragraph{Related-work generation.} 



We evaluate related-work generation on 64 target manuscripts from the OARelatedWork test split~\citep{docekal2024oarelatedwork}. Each instance contains a focal manuscript and a fixed reference set of at least five papers. Its original related-work section is withheld from the generation input. The reference set is fixed, so no additional retrieval is performed. We compare the proposed ScholarStack with the Full-text baseline. Both use the same model and \texttt{write-related-work} skill. Full-text applies this skill to the complete reference texts, whereas ScholarStack applies the same skill to frozen paper-level assets and the L1/L2/L3 knowledge layers without access to those full texts. The focal-manuscript context, citation registry, writing procedure, and 300--600-word limit are otherwise identical, so the comparison isolates the evidence representation used by the shared writing skill. An independent LLM judge evaluates each anonymized section separately on Relevance, Coverage, Specificity, Support, and Overall quality. We also audit claims against the original references as True, Wrong Citation, Unverifiable, or False, and compute citation F1 against the benchmark reference set. Appendix~\ref{app:related-work} gives selection, access, and evaluation details.
\vspace{-0.3em}
\paragraph{Literature review generation.}
We evaluate on the 25-task ReportBench-ML subset used by Lacuna~\citep{li2025reportbench,weiss2026lacuna}. Each task requests an English survey under topic, coverage, and publication-date constraints. For each task, ScholarStack uses an agentic search workflow to retrieve 100 relevant papers, compiles them into an L1--L3 knowledge view, and applies the review skill to produce the survey. Claude Code DeepResearch with Opus 4.8 model serves as the local comparator and cites 13 papers per task on average. Expert-survey content and gold reference membership are withheld from both workflows. We report micro citation precision, recall, and F1 over papers cited in the final body; the count-control condition retains the first 13 papers cited by ScholarStack in retrieval order, without gold-based reranking or regeneration. Content quality is assessed with point-wise RACE~\citep{du2025deepresearchbench} using an independent judge and shared task-specific rubrics. Published Lacuna values provide external context rather than rerun baselines. Appendix~\ref{app:review-protocol} reports resource accounting and trace-backed examples.


\vspace{-0.3em}
\paragraph{Experimental-design generation.}
\label{sec:exp-expdesign}
We construct a cross-domain benchmark for experimental-design generation from a citation-linked scientific corpus spanning 12 fields with 38 targets. Each target is a paper whose source text reports an experimental study, and the ground truth is the set of experiments that paper actually performed, extracted from its source text; each target is accompanied by a corpus of related papers (median 14 per case) that a system may draw on when designing experiments. All methods share the base model and tool budget and are closed-book, differing only in knowledge access: Full-text (w/o Skill) reads the raw files with no task method, Full-text (w/ Skill) applies our domain-agnostic design method over the same raw files, and ScholarStack applies that method over assets compiled offline from the related corpus. We report (i) a \emph{coverage F1} against the paper's actual experiments, weighted by experiment kind (main comparison/ablation $1.0$; analysis, robustness, human evaluation, theory $0.6$; other $0.3$), and (ii) a \emph{blind pairwise preference} in which a neutral LLM judge (GPT-6), shown the contribution and two anonymized designs in randomized order with an anti-verbosity instruction, selects the better design. Per-query cost is reported in tokens. Appendix~\ref{app:expdesign} details the metric, cost accounting, and honest-comparison notes.

\paragraph{Idea generation.}
\label{sec:exp-idea-generation}
We evaluate 35 cases across five broad domains drawn from IdeaBench~\citep{guo2024ideabench}, MOOSE-Chem~\citep{yang2024moosechem}, and AI Idea Bench 2025~\citep{qiu2025aiideabench}. Each case provides a research task and reference papers. Full-text (w/o Skill) generates ideas directly from the task and reference full texts; Full-text (w/ Skill) applies the idea-generation skill described in Section~\ref{sec:m-idea-generation} to the same full texts; ScholarStack retains that skill while accessing assets compiled from the same references, without directly reading the full papers. An LLM judge ranks anonymized outputs across five evaluation dimensions adopted from \citet{li2024chainofideas}: novelty, significance, clarity, feasibility, and effectiveness. First, second, and third place receive 1, 0.5, and 0, respectively; scores are averaged across cases and then equally across dimensions. Appendix~\ref{app:idea-generation-evaluation} details the benchmark composition and judging protocol.

\subsubsection{Claim and Consistency Assessment}
\label{sec:setup-verifier-claim}

\paragraph{Scientific claim verification.}
We evaluate $80$ claim--paper pairs from the public training/development split of NLPCC 2026 Shared Task 10, Track 2~\citep{nlp2ct2026task10}, balanced across \emph{Supported}, \emph{Overstate}, \emph{Topical Match}, and \emph{Irrelevant} ($20$ per label). Full-text receives the dataset-provided article text, whereas ScholarStack receives the original L1 objects compiled from the same paper. Neither condition receives task images or external retrieval. We report exact-label accuracy and mean tokens per query, including retries; failures and abstentions remain in the accuracy denominator. Evidence localization is not evaluated. Appendix~\ref{app:verifier-claim} provides selection criteria, label definitions, and cost accounting.




\subsubsection{Continuous Research}
\label{sec:continuous-setup}
We evaluate research chains on five topics over 500 QA papers from LitSearch. Each topic has an initial question and a distinct follow-up question building on its findings; both follow five stages: fixed-corpus retrieval, literature review generation, idea generation, novelty assessment, and experimental-design generation. Context retains conversation history; Notes uses fresh sessions with self-maintained notes; ScholarStack uses fresh sessions with task outputs and persistent Knowledge Views. All conditions can inspect source sentences. We assess task quality, understanding reuse, and token costs using anonymous LLM judging, rule checks, and access logs. Appendices~\ref{app:continuous-questions} and~\ref{app:continuous-metrics} provide the questions and metric definitions, respectively.

\subsection{Main Results}

\subsubsection{Literature Retrieval}

\paragraph{Fixed-corpus retrieval.}
Table~\ref{tab:retrieval-qwen} reports fixed-corpus retrieval results over all 50 LitSearch queries. On the 50 LitSearch queries, ScholarStack exceeds the Full-text baseline at R@1, R@5, and MRR@40, while the Full-text baseline is higher at R@10. ScholarStack also uses slightly less time and fewer tokens per query.
The coverage split in Table~\ref{tab:retrieval-coverage} localizes the difference. For in-source queries, ScholarStack improves all four retrieval metrics. For out-of-source queries, it ties the Full-text baseline at R@1 and R@5 while being slightly lower at R@10 and MRR@40.
These results indicate that the aggregate head-ranking gains are concentrated among queries whose gold set overlaps the papers used to compile the layer.

\begin{table}[h]
\centering
\small
\caption{Fixed-corpus literature search results on LitSearch~\citep{ajith2024litsearch}. Tokens are per-query means. Corpus and asset size refer to the number of papers. Best results are marked in \textbf{Bold}.}
\label{tab:retrieval-qwen}
\begin{tabular}{lccccccc}
\toprule
Method & Corpus & Asset Size & R@1 & R@5 & R@10 & MRR@40 & Tokens \\
\midrule
BM25 & 64,183 & -- & 0.220 & 0.470 & 0.537 & 0.331 & -- \\
Dense & 64,183 & -- & 0.257 & 0.503 & 0.533 & 0.379 & -- \\
Hybrid & 64,183 & -- & 0.347 & 0.537 & 0.583 & 0.452 & -- \\
Full-text & 64,183 & 900 & 0.650 & 0.767 & \textbf{0.787} & 0.707 & 49.3k \\
ScholarStack & 64,183 & 900 & \textbf{0.670} & \textbf{0.773} & 0.773 & \textbf{0.717} & 47.9k \\
\bottomrule
\end{tabular}%
\end{table}


\begin{table}[h]
\centering
\small
\caption{Fixed-corpus literature search results by knowledge-source coverage: 19 queries with at least one gold paper among the 900 asset sources (in-source), and 31 with none (out-of-source). All gold papers are evaluated against the same retrieval corpus. \textbf{Bold} marks the best value within each metric and coverage group, including ties.}
\label{tab:retrieval-coverage}
\begin{tabular}{lcccccccc}
\toprule
& \multicolumn{4}{c}{In-source} & \multicolumn{4}{c}{Out-of-source} \\
\cmidrule(lr){2-5}\cmidrule(lr){6-9}
Method & R@1 & R@5 & R@10 & MRR@40 & R@1 & R@5 & R@10 & MRR@40 \\
\midrule
Full-text & 0.711 & 0.807 & 0.807 & 0.781 & \textbf{0.613} & \textbf{0.742} & \textbf{0.774} & \textbf{0.662} \\
ScholarStack & \textbf{0.763} & \textbf{0.825} & \textbf{0.825} & \textbf{0.818} & \textbf{0.613} & \textbf{0.742} & 0.742 & 0.654 \\
\bottomrule
\end{tabular}%
\end{table}

\paragraph{Open-world retrieval.}
Table~\ref{tab:open-world-retrieval} reports pooled Tier~1 results over all 89 queries. Relative to matched raw-text access, ScholarStack raises precision from 0.1376 to 0.1579 (14.7\% relative) and recall from 0.1240 to 0.1338 (7.9\%), yielding a 10.7\% improvement in F1. Macro and Micro Weighted Recall increase by 6.6\% and 11.5\%, respectively. The source papers and online access budget are matched: Full-text can inspect the same papers in full text, whereas ScholarStack exposes their precompiled cross-layer representation. The difference is therefore associated with representation and access rather than additional source documents. The experiment evaluates the layer as a whole and does not attribute the gain separately to L1, L2, or L3.

\begin{table}[h]
\centering
\small
\caption{Pooled open-world retrieval results on 89 SAGE queries. All metrics use the explicit Tier~1 set, which contains approximately five papers. Recall, Precision, F1, and Macro Weighted Recall are macro-averaged over queries; Micro Weighted Recall pools weighted labels.}
\label{tab:open-world-retrieval}
\begin{tabular}{lccccc}
\toprule
Method & Recall & Precision & F1-Score & Macro W-Recall & Micro W-Recall \\
\midrule
Full-text & 0.124 & 0.138 & 0.128 & 0.117 & 0.113 \\
ScholarStack & \textbf{0.134} & \textbf{0.158} & \textbf{0.141} & \textbf{0.124} & \textbf{0.126} \\
\bottomrule
\end{tabular}
\end{table}


\paragraph{Novelty assessment.}
Table~\ref{tab:novelty} reports novelty assessment results across 12 fields. ScholarStack uses 21.4k tokens per query, compared with 35.6k for the Full-text method (about $40\%$ fewer). Under our decision-relevance axis weighting its weighted identification F1 is also higher, verifying the benefit of structured knowledge layers.

\begin{table}[htbp]
\centering\small
\caption{Novelty-assessment results for $n{=}40$ overlap-present targets across 12 fields and $135$ gold overlaps. Weighted identification F1 uses weights of $1.0$ for core axes and $0.3$ for peripheral axes; token counts are per-query means.}
\label{tab:novelty}
\begin{tabular}{lcccc}
\toprule
Method & Weighted F1 & Recall & Precision & Tokens \\
\midrule
Full-text                  & 0.645 & 0.761 & 0.560 & 35.6k \\
\textbf{ScholarStack} & \textbf{0.662} & \textbf{0.779} & \textbf{0.576} & \textbf{21.4k} \\
\bottomrule
\end{tabular}
\end{table}

\subsubsection{Scientific Question Answering}
\label{sec:scientific-qa}

\paragraph{Single-paper QA.}
Tables~\ref{tab:qasper-main} and~\ref{tab:peerqa-main} report results on two single-paper QA benchmarks. On QASPER, ScholarStack improves overall Answer-F1 (0.571 versus 0.561) and Evidence-F1 (0.589 versus 0.561) over RAG while using 12\% fewer input tokens. ScholarStack reaches about 93\% of the full-text Answer-F1 at 34\% of its input tokens.

\begin{table}[htbp]
\centering
\small
\caption{Single-paper QA on QASPER (416 papers / 1{,}428 questions). Overall denotes Answer-F1 across all questions; Extr., Abst., Bool., and Unans. denote extractive, abstractive, boolean, and unanswerable questions, respectively. Answer-F1 measures token-level answer overlap; Evidence-F1 measures paragraph-level evidence overlap. Human is the reported human-answer reference from QASPER~\citep{dasigi-etal-2021-dataset}. Tokens denotes mean input tokens per question; \textbf{bold} marks improvements over RAG.}
\label{tab:qasper-main}
\begin{tabular}{lccccccc}
\toprule
 & \multicolumn{5}{c}{Answer-F1} & & \\
\cmidrule(lr){2-6}
Method & Overall & Extr. & Abst. & Bool. & Unans. & Evidence-F1 & Tokens \\
\midrule
Full-text    & 0.615 & 0.660 & 0.270 & 0.695 & 0.920 & 0.602 & 5.4k \\
Human        & 0.609 & --    & --    & --    & --    & 0.716 & --   \\
\midrule
RAG          & 0.561 & 0.579 & 0.253 & 0.657 & 0.909 & 0.561 & 2.1k \\
ScholarStack & \textbf{0.571} & \textbf{0.599} & \textbf{0.263} & 0.650 & \textbf{0.911} & \textbf{0.589} & \textbf{1.9k} \\
\bottomrule
\end{tabular}%
\end{table}

On PeerQA, ScholarStack improves all five aggregate answerability metrics and free-form ROUGE-L over RAG while using 28\% fewer input tokens. ScholarStack reaches about 93\% of the full-text Macro-F1 with about 11\% of its input tokens. Appendix~\ref{app:single-qa} reports domain-level results and selected qualitative examples.

\begin{table}[htbp]
\centering
\small
\caption{Single-paper QA on PeerQA (208 papers / 579 questions). Answerability is evaluated on 495 questions: Ans-F1 and Unans-F1 denote F1 for answerable and unanswerable questions; Acc., W-F1, and Macro-F1 denote accuracy, class-frequency-weighted F1, and unweighted mean class F1. Generation ROUGE-L is evaluated on 245 questions against free-form reference answers (FF) and annotated evidence (AE). Human denotes model answers using human-annotated evidence, with answerability scored only on answerable questions~\citep{baumgartner-etal-2025-peerqa}. Tokens denotes mean input tokens per question; \textbf{bold} marks improvements over RAG.}
\label{tab:peerqa-main}
\begin{tabular}{lccccccc c}
\toprule
 & \multicolumn{5}{c}{Answerability} & \multicolumn{2}{c}{Generation (Rouge-L)} & \\
\cmidrule(lr){2-6}\cmidrule(lr){7-8}
Method & Ans-F1 & Unans-F1 & Acc. & W-F1 & Macro-F1 & FF & AE & Tokens \\
\midrule
Full-text    & 0.806 & 0.507 & 0.721 & 0.738 & 0.656 & 0.155 & 0.205 & 12.6k \\
Human        & 0.732 & --    & --    & --    & --    & 0.150 & 0.161 & 0.3k   \\
\midrule
RAG          & 0.736 & 0.476 & 0.649 & 0.677 & 0.606 & 0.142 & 0.156 & 1.9k  \\
ScholarStack & \textbf{0.749} & \textbf{0.478} & \textbf{0.661} & \textbf{0.687} & \textbf{0.613} & \textbf{0.145} & 0.156 & \textbf{1.4k} \\
\bottomrule
\end{tabular}%
\end{table}
\paragraph{Multi-paper QA.}
Table~\ref{tab:mdaqa-main} reports results on 797 MDAQA questions. With the same answering model, ScholarStack achieves an overall score of 18.41, compared with 8.46 for Full-text, an absolute gain of 9.95. Compared with Full-text, ScholarStack improves correctness from 1.11 to 3.89, completeness from 2.01 to 3.35, cross-paper synthesis from 1.41 to 3.48, calibration from 0.92 to 3.90, and directness from 3.01 to 3.80. The improvements in synthesis and calibration suggest that ScholarStack organizes evidence in a form that better supports cross-paper integration and evidence-aligned answers under the same answering model. At the same time, ScholarStack reduces answer-generation costs, using 5.5k total tokens per question compared with 20.8k for Full-text, a reduction of approximately 73.3\%. The mean generation time also decreases from 13.13 to 7.54 seconds per question.


\begin{table}[htbp]
\centering\small
\caption{Multi-paper QA on 797 MDAQA questions. Corr., Comp., Synth., Cal., and Dir. denote correctness, completeness, cross-paper synthesis, calibration, and directness. Scores are from 0 to 4, and Total sums the five scores (0 to 20). Tokens and Sec. are mean tokens and generation seconds per question over available usage records, excluding knowledge construction and judging.}
\label{tab:mdaqa-main}
\begin{tabular}{lcccccccc}
\toprule
Method & Corr. & Comp. & Synth. & Cal. & Dir. & Total & Tokens & Sec. \\
\midrule
Full-text & 1.11 & 2.01 & 1.41 & 0.92 & 3.01 & 8.46 & 20.8k & 13.13 \\
ScholarStack & \textbf{3.89} & \textbf{3.35} & \textbf{3.48} & \textbf{3.90} & \textbf{3.80} & \textbf{18.41} & \textbf{5.5k} & \textbf{7.54} \\
\bottomrule
\end{tabular}
\end{table}

ScholarStack achieves higher completeness and synthesis scores while using shorter inputs. The knowledge layers organize information at the levels needed for cross-paper answering. The representations of L1, L2 and L3 establish together an explicit structure for comparing findings and tracing conclusions back to their sources. In the evaluation, this organization is associated with higher coverage and better calibration, while requiring fewer generation tokens and less recorded generation time. This combination is particularly useful for multi-paper questions, where an answer must integrate several findings while preserving the conditions under which each applies. The appendix reports the results for the intermediate-level L1 and L1+L2 configurations, as well as those using the Cross-Paper Evidence Synthesis (CPES) skill.

\subsubsection{Evidence-Grounded Generation}
\paragraph{Related-work generation.}

Table~\ref{tab:related-work-quality} shows that ScholarStack outperforms Full-text on four of the five independently judged quality dimensions. Its gains in Coverage ($4.250$ vs.\ $4.219$), Specificity ($4.453$ vs.\ $4.359$), Support ($4.438$ vs.\ $4.234$), and Overall quality ($4.203$ vs.\ $4.156$) indicate that the knowledge layer helps the writer cover the relevant literature, make more concrete comparisons, and ground the resulting synthesis in evidence. Table~\ref{tab:related-work-factuality} shows the same favorable pattern in evidence use. ScholarStack increases the True rate from $91.82\%$ to $92.43\%$, while reducing Wrong Citation from $2.14\%$ to $1.10\%$ and False claims from $1.81\%$ to $1.15\%$. It also improves citation F1 from $0.95$ to $0.97$, showing closer agreement with the references selected by the target authors. Although its Unverifiable rate is slightly higher, ScholarStack produces fewer citation errors and false claims while selecting a more appropriate reference set.

\begin{table}[htbp]
\centering\small
\caption{Quality-dimension performance on the OARelatedWork benchmark (each dimension scored out of 5). Rel., Cov., Spec., and Supp. denote Relevance, Coverage, Specificity, and Support.}
\label{tab:related-work-quality}
\begin{tabular}{lccccc}
\toprule
Method & Rel. & Cov. & Spec. & Supp. & Overall \\
\midrule
Full-text    & \textbf{4.688} & 4.219 & 4.359 & 4.234 & 4.156 \\
{ScholarStack} & 4.578 & \textbf{4.250} & \textbf{4.453} & \textbf{4.438} & \textbf{4.203} \\
\bottomrule
\end{tabular}
\end{table}

\begin{table}[htbp]
\centering\small
\caption{Factuality and citation performance on the OARelatedWork benchmark. Claim-label values are sample-mean percentages; citation F1 uses a 0--1 scale.}
\label{tab:related-work-factuality}
\begin{tabular}{lccccc}
\toprule
Method & True $\uparrow$ & \shortstack{Wrong Cit. $\downarrow$} & \shortstack{Unverif. $\downarrow$} & False $\downarrow$ & Citation F1 $\uparrow$ \\
\midrule
Full-text    & 91.82 & 2.14 & \textbf{4.23} & 1.81 & 0.95 \\
{ScholarStack} & \textbf{92.43} & \textbf{1.10} & 5.31 & \textbf{1.15} & \textbf{0.97} \\
\bottomrule
\end{tabular}
\end{table}

Taken together, ScholarStack outperforms Full-text on most quality dimensions and on the key citation and factuality indicators. L2 organization and L3 scoped syntheses expose coherent research threads and cross-paper relations, while L1 facts retain the concrete source-grounded details needed to substantiate them. This layered representation lets the writer work from reusable structured assets instead of repeatedly processing full papers, while producing more comprehensive, specific, and well-supported related work with better citation selection.

\paragraph{Literature review generation.}

Table~\ref{tab:ml25-citation} shows that ScholarStack achieves the strongest reference coverage, recovering 637 expert-survey references and reaching 0.182 recall and 0.214 micro F1. Claude Code recovers 171 references, with 0.049 recall and 0.090 F1. ScholarStack's lower full-set precision (0.260 versus 0.553) follows from its substantially broader search: the Agentic Search workflow retrieves approximately 100 papers for each survey, enabling the final report to cover more relevant directions but also increasing the number of citations outside the expert reference set.

To compare the systems at a similar citation budget, we restrict ScholarStack to the first 13 papers it cites in retrieval order, matching Claude Code's average. Under this control, ScholarStack reaches 0.603 precision, 0.056 recall, and 0.103 F1, exceeding Claude Code on all three metrics. The result shows that the coverage gain is not obtained only by adding a long tail of references: the early ScholarStack citation set is itself more concentrated on expert-survey references, while the remaining citations provide the additional breadth observed in the full condition.

Table~\ref{tab:ml25-race} further shows that ScholarStack achieves the highest report-quality score overall (9.18) and in every RACE dimension. The largest gains over Claude Code are in comprehensiveness (9.23 versus 7.04) and insight (9.30 versus 7.43), consistent with L2 providing a corpus-level organization and L3 connecting findings across papers. L1 evidence and the review skill's source-checking workflow help preserve technical conditions and support coherent, well-grounded exposition, contributing to the gains in instruction following and readability as well. These scores evaluate the complete workflows; Appendix~\ref{app:review-cases} provides trace-backed examples of the three mechanisms and discusses the evaluation boundaries in detail.

\begin{table}[htbp]
\centering
\small
\caption{ReportBench-ML citation overlap against expert-survey references.}
\label{tab:ml25-citation}
\setlength{\tabcolsep}{5pt}
\begin{tabular}{lcccc}
\toprule
System & F1 & Prec. & Rec. & Hits \\
\midrule
GPT-Researcher & 0.039 & 0.290 & 0.022 & 72 \\
STORM & 0.015 & 0.250 & 0.008 & 21 \\
LangChain ODR & 0.007 & 0.055 & 0.004 & 13 \\
Lacuna Deep Research & 0.052 & 0.339 & 0.028 & 99 \\
\midrule
Claude Code DR & 0.090 & 0.553 & 0.049 & 171 \\
{ScholarStack} & \textbf{0.214} & 0.260 & \textbf{0.182} & \textbf{637} \\
\quad\textit{First 13 cited} & 0.103 & \textbf{0.603} & 0.056 & 196 \\
\bottomrule
\end{tabular}
\end{table}

\begin{table}[htbp]
\centering
\small
\caption{RACE report-quality scores on ReportBench-ML. Scores are out of 10.}
\label{tab:ml25-race}
\setlength{\tabcolsep}{5pt}
\begin{tabular}{lccccc}
\toprule
System & Overall & Comp. & Insight & Instr. & Read. \\
\midrule
GPT-Researcher & 5.24 & 5.31 & 4.87 & 4.41 & 7.62 \\
STORM & 2.90 & 3.45 & 2.67 & 1.66 & 4.20 \\
LangChain ODR & 7.42 & 7.48 & 7.23 & 7.22 & 8.16 \\
Lacuna Deep Research & 7.82 & 8.01 & 7.61 & 7.57 & 8.34 \\
\midrule
Claude Code DR & 7.54 & 7.04 & 7.43 & 8.28 & 7.88 \\
{ScholarStack} & \textbf{9.18} & \textbf{9.23} & \textbf{9.30} & \textbf{9.16} & \textbf{8.81} \\
\bottomrule
\end{tabular}
\end{table}



\paragraph{Experimental-design generation.}

Table~\ref{tab:expdesign} reports coverage and cost; Table~\ref{tab:expdesign-pref} reports the blind preference. On coverage F1, ScholarStack and Full-text (w/ Skill) are essentially tied ($0.404$ vs.\ $0.398$; we do not read this $0.006$ gap as a quality difference), and both clearly exceed Full-text (w/o Skill) ($0.339$). Both methods are strongly preferred over Full-text (w/o Skill) by the blind judge ($32/38$ and $34/38$ wins); the gain is precision-driven, as Full-text (w/o Skill) over-proposes experiments (mean $22.9$ vs.\ $18.6$--$18.7$) and is penalized for padding. Head-to-head, the judge slightly favors Full-text (w/ Skill) over ScholarStack ($22$ vs.\ $16$), so we do \emph{not} claim ScholarStack improves design quality over Full-text (w/ Skill). ScholarStack's contribution is matching the quality of Full-text (w/ Skill) at the lowest per-query cost of the three methods---about $40\%$ fewer tokens than Full-text (w/ Skill)---while requiring no task-specific assets.

\begin{table}[h]
\centering\small
\caption{Experimental-design generation results. Coverage F1 against the paper's actual experiments (weighted by experiment kind). The Tokens column reports per-query token usage.}
\label{tab:expdesign}
\begin{tabular}{lcccc}
\toprule
Method & Coverage F1 & Recall & Precision & Tokens \\
\midrule
Full-text (w/o Skill) & 0.339 & \textbf{0.665} & 0.255 & 18.7k \\
Full-text (w/ Skill)  & 0.398 & \textbf{0.665} & 0.329 & 20.7k \\
ScholarStack         & \textbf{0.404} & 0.642 & \textbf{0.331} & \textbf{12.4k} \\
\bottomrule
\end{tabular}
\end{table}

\begin{table}[h]
\centering\small
\caption{Blind pairwise preference (GPT-6 judge; anonymized, randomized order, anti-verbosity instruction). Wins/losses/ties over $38$ targets.}
\label{tab:expdesign-pref}
\begin{tabular}{lccc}
\toprule
Comparison & Wins & Losses & Ties \\
\midrule
Full-text (w/ Skill) vs.\ Full-text (w/o Skill) & \textbf{32} & 6 & 0 \\
ScholarStack vs.\ Full-text (w/o Skill)        & \textbf{34} & 4 & 0 \\
ScholarStack vs.\ Full-text (w/ Skill)         & 16 & 22 & 0 \\
\bottomrule
\end{tabular}
\end{table}

\paragraph{Idea generation.}
Table~\ref{tab:idea-generation} shows that ScholarStack achieves the highest score on all five dimensions, with an overall mean of 0.577, compared with 0.526 for Full-text (w/ Skill) and 0.397 for Full-text (w/o Skill). The generation method improves the mean score by 0.129 under full-text access, while the knowledge layer adds a further 0.051 with the method held fixed.

\begin{table}[h]
\centering
\caption{Comparison of idea-generation quality across five dimensions on the benchmark.}
\label{tab:idea-generation}
\small
\setlength{\tabcolsep}{5pt}
\begin{tabular}{lcccccc}
\toprule
Method & Novelty & Significance & Clarity & Feasibility & Effectiveness & Mean \\
\midrule
Full-text (w/o Skill) & 0.486 & 0.457 & 0.343 & 0.257 & 0.443 & 0.397 \\
Full-text (w/ Skill) & 0.500 & 0.514 & 0.571 & 0.543 & 0.500 & 0.526 \\
{ScholarStack} & \textbf{0.514} & \textbf{0.529} & \textbf{0.586} & \textbf{0.700} & \textbf{0.557} & \textbf{0.577} \\
\bottomrule
\end{tabular}
\end{table}

The largest gains of Full-text (w/ Skill) over Full-text (w/o Skill) occur in clarity and feasibility. This pattern is consistent with the generation method's explicit organization of candidate ideas and its emphasis on resource constraints and internal refinement. The gains of ScholarStack over Full-text (w/ Skill) suggest that the structured organization of reference evidence supports idea development and refinement. Preserved applicability conditions inform feasibility judgments, while relations among prior approaches support novelty by facilitating the identification of new combinations and distinct contributions. Cross-study syntheses of limitations and conditional findings provide an evidential basis for explaining how proposed directions could improve on existing work, consistent with the gain in effectiveness.

\subsubsection{Claim and Consistency Assessment}
\label{sec:exp-verifier-claim}

\paragraph{Scientific claim verification.}
Table~\ref{tab:verifier-claim} compares the two paper contexts on all $80$ claim--paper pairs. ScholarStack yields $41/80$ correct predictions versus $37/80$ for the Full-text baseline, an observed accuracy gain of $5.00$ percentage points. Mean usage falls from $12{,}046.29$ to $2{,}817.10$ tokens per query, a reduction of $76.61\%$. The L1 view reduces query-time token use while retaining useful support-relation information. This single-run accuracy comparison on a small, balanced subset is descriptive, with no significance test and nearly half of the labels incorrect. Reliability and total cost, including construction, require further evaluation.

\begin{table}[h]
\centering\small
\caption{Scientific claim verification on $80$ NLPCC instances ($20$ per label). Tokens/query includes input, output, and retries, excluding knowledge construction.}
\label{tab:verifier-claim}
\begin{tabular}{lrrr}
\toprule
Method & Correct & Accuracy (\%) & Tokens/query \\
\midrule
Full-text & 37/80 & 46.25 & 12.0k \\
ScholarStack & \textbf{41/80} & \textbf{51.25} & \textbf{2.8k} \\
\bottomrule
\end{tabular}
\end{table}

\subsubsection{Continuous Research and Understanding Reuse}
\label{sec:continuous-research}

We further evaluate whether understanding developed during research can support subsequent tasks and follow-up questions in a continuous chain. We compare three ways of retaining research understanding across tasks: \textbf{Context}, which carries forward the conversation history; \textbf{Notes}, which passes self-maintained notes between fresh sessions; and \textbf{ScholarStack}, which reuses persistent research assets through Knowledge Views.

Table~\ref{tab:continuous-main} shows that ScholarStack achieves strong performance across all five tasks. In literature review generation, ScholarStack achieves citation coverage of 0.720, exceeding Context and Notes by 12.7 and 33.0 percentage points, respectively, and also achieves the highest RACE score. In novelty assessment, ScholarStack achieves overlap support of 0.925, exceeding Context by 12.5 and Notes by 25.0 percentage points. ScholarStack also achieves the highest retrieval recall and matches the best scores for idea and experimental-design generation. Across the whole chain, it passes all checklist items for task requirements, compared with compliance rates of 0.953 for Context and 0.988 for Notes. ScholarStack uses 59\% fewer new tokens than Context and 9\% fewer than Notes. Chain costs include query, generation, and write-back/checking, but exclude initial knowledge-layer construction.

\begin{table}[htbp]
\centering\small
\caption{Task quality, chain-level compliance, and token cost, averaged over five topics. Citation coverage measures coverage of relevant papers. Review quality follows the RACE evaluation used in the preceding literature review task. Overlap support measures evidence support for novelty-assessment judgments. New tokens include non-cached input and output, excluding knowledge-layer construction. Metric definitions are provided in Appendix~\ref{app:continuous-metrics}. \textbf{Bold} indicates the best results in each metric. Higher is better ($\uparrow$); lower is better ($\downarrow$).}
\label{tab:continuous-main}
\setlength{\tabcolsep}{4pt}
\begin{tabular}{@{}llrrr@{}}
\toprule
Task & Metric & Context & Notes & ScholarStack \\
\midrule
Fixed-corpus retrieval & Recall@100 $\uparrow$ & 0.819 & 0.772 & \textbf{0.832} \\
Literature review generation & Citation coverage $\uparrow$ & 0.593 & 0.390 & \textbf{0.720} \\
 & RACE overall (0--10) $\uparrow$ & 6.80 & 5.60 & \textbf{7.00} \\
Idea generation & Quality (1--5) $\uparrow$ & \textbf{4.36} & 4.20 & \textbf{4.36} \\
Novelty assessment & Overlap support $\uparrow$ & 0.800 & 0.675 & \textbf{0.925} \\
Experimental-design generation & Overall score (0--4) $\uparrow$ & 2.60 & \textbf{3.20} & \textbf{3.20} \\
\midrule
Whole chain & Checklist compliance $\uparrow$ & 0.953 & 0.988 & \textbf{1.000} \\
 & New tokens (k) $\downarrow$ & 1,214 & 552 & \textbf{501} \\
\bottomrule
\end{tabular}
\end{table}

\begin{table}[htbp]
\centering\small
\caption{Core follow-up-chain results, averaged over five chains. Correct reuse coverage is the fraction of ten expected findings cited for the specified paper and aspect in follow-up deliverables (Appendix~\ref{app:continuous-metrics}). Input change includes cached input and is relative to each method's own first chain. Raw-document reads exclude queries for registered evidence sentences.}
\label{tab:continuous-reuse}
\setlength{\tabcolsep}{5pt}
\begin{tabular}{@{}lrrr@{}}
\toprule
Metric & Context & Notes & ScholarStack \\
\midrule
Correct reuse coverage $\uparrow$ & \textbf{1.000} & 0.920 & 0.980 \\
Literature review citation coverage $\uparrow$ & 0.615 & 0.438 & \textbf{0.643} \\
New tokens per chain (k) $\downarrow$ & 3,559 & 581 & \textbf{491} \\
Input change $\downarrow$ & +223\% & +1\% & \textbf{-13\%} \\
Raw-document reads per chain $\downarrow$ & 187 & 105 & \textbf{0} \\
\bottomrule
\end{tabular}
\end{table}

Table~\ref{tab:continuous-reuse} shows that ScholarStack maintains near-complete correct reuse coverage with fewer new tokens than either baseline and the highest literature-review citation coverage. Relative to each method's own first chain, total input tokens (including cached input) decrease by 13\% for ScholarStack's follow-up chain and increase by 223\% for Context's. Initial knowledge-layer construction requires approximately 23M input and 2M output tokens. Including this one-time cost, ScholarStack uses 68.3M total input and output tokens across the two chains of all five topics, compared with 389.8M for Context. Relative to Context, the construction cost is recovered within the second chain of a topic (Appendix~\ref{app:continuous-cost}). Queries in each ScholarStack follow-up chain return 21.8 distinct first-chain objects on average; L3 syntheses account for 82\% of accesses to first-chain objects, including repeated returns. These objects make individual judgments and their conditions, evidence, and revision state available to later tasks, supporting reuse without reconstructing a judgment's basis.

Table~\ref{tab:continuous-layers} examines layer contributions using fixed-query retrieval tests and single-topic replays. These settings differ from the five-topic complete-chain evaluations in Tables~\ref{tab:continuous-main} and~\ref{tab:continuous-reuse}, so absolute scores are not directly comparable. L2 improves retrieval recall, and L2/L3 broaden literature-review citation coverage. Adding L3 to L1+L2 raises novelty overlap support from 0.875 to 1.000. With upstream outputs fixed, replaying experimental-design generation under all three layer configurations retains full chain-level checklist compliance. Full-layer access also yields the highest observed follow-up reuse coverage. These results support selecting layers according to task needs. Appendix~\ref{app:continuous} gives the questions, metric definitions, and an example illustrating how later tasks reuse and revise evidence-linked judgments.

\begin{table}[!htbp]
\centering\small
\caption{Layer-access comparisons. Retrieval directly tests L1 and L1+L2 with fixed query terms across five topics, rather than the full agent workflow. Other rows use only \emph{Conversational QA}. Stage replays fix upstream snapshots; follow-up tests fix the first-chain terminal state. Object accesses include repeated returns. Retrieval does not use L3.}
\label{tab:continuous-layers}
\setlength{\tabcolsep}{4pt}
\begin{tabular}{@{}llrrr@{}}
\toprule
Task & Metric & L1 & L1+L2 & L1+L2+L3 \\
\midrule
\multicolumn{5}{@{}l}{\textit{Within-stage comparisons}} \\
Fixed-corpus retrieval & Recall@100 $\uparrow$ & 0.723 & \textbf{0.774} & --- \\
Literature review generation & Citation coverage $\uparrow$ & 0.733 & 0.867 & \textbf{0.978} \\
 & RACE overall $\uparrow$ & 8.20 & 8.60 & \textbf{8.80} \\
Novelty assessment & Overlap support $\uparrow$ & 0.875 & 0.875 & \textbf{1.000} \\
\midrule
\multicolumn{5}{@{}l}{\textit{Across-chain comparisons}} \\
Follow-up & Correct reuse coverage $\uparrow$ & 0.900 & 0.800 & \textbf{1.000} \\
& Overlap support $\uparrow$ & 0.875 & 0.750 & \textbf{1.000} \\
& First-chain object accesses $\uparrow$ & 0 & 16 & 121 \\
\bottomrule
\end{tabular}
\end{table}

\subsection{Analysis of Layer Contributions}
\label{sec:exp-layer-roles}

The results above show that compiled assets can improve both query-time efficiency and the quality of research outputs. The asset layers contribute to these improvements in different ways across tasks: some tasks rely primarily on source-grounded facts, while others also draw on domain organization or existing syntheses.

L1 preserves source statements together with their study conditions and evidence references. These records provide the basis for answering questions about a paper and assessing whether its findings support a claim. In claim verification, Table~\ref{tab:verifier-claim} shows that using L1 alone reduces token use relative to Full-text, with a small observed improvement in accuracy. Beyond these paper-specific tasks, L1 provides evidence for identifying substantive overlaps in novelty assessment and tracing claims to their sources in scholarly writing. For idea generation and experimental design, reported findings, procedures, and evaluation conditions supply concrete premises for assessing proposed directions and choosing appropriate tests.

L2 organizes research objects through shared identities, categories, and relations. In fixed-corpus retrieval, this structure guides candidate selection; in open-world retrieval, it helps formulate search terms and identify related approaches to investigate. This role in candidate selection is supported by the retrieval results from continuous research reported in Table~\ref{tab:continuous-layers}: adding L2 to L1 raises Recall@100 from 0.723 to 0.774 using matched query terms across five topics. Beyond retrieval, categories group studies into research threads for related-work and review generation, while taxonomy dimensions provide explicit axes for comparison.

L3 provides scoped interpretations of findings across studies. These syntheses support multi-paper answers, position contributions in relation to prior work, and explain methodological developments in surveys. Table~\ref{tab:mdaqa-ablation-cost} provides evidence from MDAQA: with the base evidence package and Bare answering procedure held fixed, adding L3 to L1+L2 increases the total score from 17.01 to 18.41 and the cross-paper synthesis score from 2.80 to 3.48. In Table~\ref{tab:continuous-layers}, the literature-review results from continuous research show a similar benefit from L3: adding L3 to L1+L2 increases citation coverage from 0.867 to 0.978 and the RACE overall score from 8.60 to 8.80. These observations are consistent with scoped syntheses helping tasks integrate relevant studies into a broader account. The continual-learning review in Appendix~\ref{app:review-cases} illustrates this interpretive role in writing by connecting successive methods through the limitations they address. For idea generation, such accounts can reveal unresolved questions and motivate directions for further investigation. In continuous research, retaining these interpretations allows later tasks to revisit judgments formed earlier. In the follow-up comparison in Table~\ref{tab:continuous-layers}, conducted from the same first-chain state, queries with full-layer access return first-chain objects 121 times, compared with 16 under L1+L2. The revision trace in Appendix~\ref{app:continuous-example} shows how counterevidence encountered during novelty assessment narrows an earlier review synthesis, providing a revised premise for experimental design.

Across these tasks, the assets support reuse of both reported content and the understanding derived from it. Tasks draw on source facts, domain organization, and scoped syntheses according to the evidence and reasoning they require.

\section{Conclusion}

We presented ScholarStack, a layered research asset framework for literature-based scientific agents. It addresses the task-specific research bottleneck by separating source-grounded facts, domain organization, and scoped cross-paper syntheses while linking them through shared identifiers, evidence references, and versioning. Task-adapted knowledge views make these assets reusable across retrieval, question answering, evidence-grounded generation, and claim assessment, while retaining their study conditions and evidential scope.

Our experiments show that structured assets can improve task effectiveness and reduce query-time token use, with benefits varying by task and source coverage. 
On SAGE, ScholarStack improves Tier 1 retrieval F1 by 10.7\% compared to the full-text baseline. Reusing assets built for novelty assessment also yields comparable experimental-design coverage F1 with about 40\% fewer query-time tokens. 
Continuous research chains further demonstrate that earlier findings and scoped judgments can be retrieved and revised in later work, with lower per-chain token use than continuous-context and persistent-note baselines. These query-time savings exclude initial asset construction costs, which can be amortized with repeated reuse. In general, ScholarStack offers a foundation for cumulative scientific assistance in which research knowledge remains inspectable and revisable across tasks. Future work will examine longer task sequences and evolving corpora, strengthen evidence verification, and account systematically for construction, maintenance, and reuse costs. 

\newpage
\section{Contributions}
The names are listed in alphabetical order by last name.

\textbf{Leader}

Wotao Yin

\textbf{Core Contributors}

Caoqinwei Gong, Xue Jiang, Wei Luo, Xiaoyu Qiu, Jiayi Sheng, Yi Wang, Zheng Yu, Ao Zhang, Haifan Zhang, Hanwei Zhang, Jihai Zhang

\textbf{Contributors}

Yuan Cao, Wei Chen, Liyun Dai, Wenkai Fang, Guanglei Wang, Kai Ying, Tingyu Zhu

\bibliographystyle{plainnat}
\bibliography{references}

\appendix
\clearpage
\section{Case Study of Layered Knowledge Construction}
\label{app:cross-layer-example}

This appendix illustrates the knowledge representation and construction described in Section~\ref{sec:representation}, using variance reduction in stochastic optimization as a single research topic. It follows source sentences through L1 facts and L2 domain organization to an L3 scoped synthesis. SAGA~\citep{defazio2014saga} is the focal paper; Prox-SVRG~\citep{xiao2014proxsvrg} and a study of saddle-point methods~\citep{palaniappan2016saddle} provide the cross-paper evidence. Figure~\ref{fig:app-topic-flow} gives the reading path, and Table~\ref{tab:app-object-fields} summarizes the objects used below.

\begingroup
\definecolor{klblue}{RGB}{39,76,119}
\definecolor{klgreen}{RGB}{42,108,91}
\definecolor{klpurple}{RGB}{105,77,137}
\newcommand{\klcard}[3]{%
  \par\smallskip\noindent\fcolorbox{#1!65}{#1!3}{%
  \begin{minipage}{\dimexpr\linewidth-2\fboxsep-2\fboxrule\relax}
  \small\raggedright\textbf{\textcolor{#1}{#2}}\par\smallskip #3
  \end{minipage}}\par\smallskip}

\begin{figure}[htbp]
\centering
\begin{tikzpicture}[x=1cm,y=1cm,
  every node/.style={font=\small,align=center},
  stage/.style={draw,rounded corners=3pt,text width=3.45cm,minimum height=2.15cm,inner sep=5pt},
  flow/.style={->,thick,>=stealth},
  trace/.style={->,dashed,>=stealth,draw=black!65}]
\node[draw=black!40,rounded corners=3pt,text width=11.8cm,inner sep=7pt] (papers) at (0,2.65)
  {\textbf{Topic: variance reduction in stochastic optimization}\\
   SAGA \quad $\cdot$ \quad Prox-SVRG \quad $\cdot$ \quad Saddle-point methods};
\node[stage,draw=klblue,fill=klblue!5] (l1) at (-4.25,0)
  {\textbf{L1:}\\\textbf{Scientific facts}\\[3pt]Method descriptions\\Conditions and guarantees\\Source sentences};
\node[stage,draw=klgreen,fill=klgreen!5] (l2) at (0,0)
  {\textbf{L2:}\\\textbf{Domain organization}\\[3pt]Shared method identity\\Family / guarantee labels\\Evaluation relations};
\node[stage,draw=klpurple,fill=klpurple!5] (l3) at (4.25,0)
  {\textbf{L3:}\\\textbf{Scoped synthesis}\\[3pt]Cross-paper account\\Declared coverage\\Supporting / limiting basis};
\draw[flow] (papers.south -| l1.north) -- node[left,font=\footnotesize]{Extract facts} (l1.north);
\draw[flow] ([xshift=1cm]l1.north) -- (-3.25,1.45) -- node[above,font=\footnotesize]{Organize facts} (-1,1.45) -- ([xshift=-1cm]l2.north);
\draw[flow] ([xshift=1cm]l2.north) -- (1,1.45) -- node[above,font=\footnotesize]{Build synthesis} (3.25,1.45) -- ([xshift=-1cm]l3.north);
\draw[trace] (l2.south) -- (0,-1.45) -- node[below,font=\footnotesize]{assignment basis} (-3.7,-1.45) -- (-3.7,-1.075);
\draw[trace] (l3.south) -- (4.25,-2.05) -- node[below,font=\footnotesize]{synthesis basis} (-4.7,-2.05) -- (-4.7,-1.075);
\draw[trace] (l1.west) -- (-6.5,0) -- node[left,rotate=90,anchor=south,font=\footnotesize]{source lookup} (-6.5,2.65) -- (papers.west);
\end{tikzpicture}
\caption{Construction and evidence access within one topic. Solid arrows show source-to-L1 extraction, L1-to-L2 organization, and L2-guided evidence selection for L3 synthesis. Dashed arrows show L2 assignments and the L3 synthesis citing L1 facts, followed by L1 source lookup through sentence maps. These are selected links in the case study, not an executed workflow or a complete dependency graph.}
\label{fig:app-topic-flow}
\end{figure}

\begin{table}[htbp]
\centering\small
\caption{Object fields used in the example. These are presentation fields for the representation in Section~\ref{sec:representation}, not a complete storage schema. All objects have an identity; verification and lifecycle states are annotations. The final column gives illustrative types and values, not exhaustive enumerations.}
\label{tab:app-object-fields}
\renewcommand{\arraystretch}{1.2}
\begin{tabular}{@{}>{\raggedright\arraybackslash}p{.20\linewidth}>{\raggedright\arraybackslash}p{.38\linewidth}>{\raggedright\arraybackslash}p{.36\linewidth}@{}}
\toprule
Object & Core fields & Example types or values \\
\midrule
Scientific Fact & statement, context, evidence & Method description; theoretical result; limitation \\
Domain Taxonomy & dimensions, categories, membership criteria & Method family; assumption regime; guarantee type \\
Canonical Entity & name, type, aliases, source anchors & Method: SAGA; dataset: MNIST \\
Category Assignment & subject, dimension, category, basis, inference type, certainty & SAGA $\in$ variance reduction; system judgment \\
Typed Relation & source, relation, target, evidence, payload & Paper evaluated on dataset; reported metric \\
Scoped Synthesis & account, scope, supporting / opposing / limiting basis & Method-family overview; selected studies \\
\bottomrule
\end{tabular}
\end{table}

\paragraph{How to read the example.}
Names such as \texttt{saga.method} are display aliases, not store identifiers. A locator such as \texttt{SAGA:20} means sentence 20 in the saved source map and does not number the example objects. Bibliographic citations identify the papers; prefixes \texttt{SAGA}, \texttt{Prox-SVRG}, and \texttt{Saddle} identify the three sources. Cards abridge saved records; locators that a stored record does not carry directly were resolved from the saved maps. The source card reproduces selected sentences, and the L3 card is an editorially revised illustration.

\subsection{L1: From a Paper to Scientific Facts}
\label{app:l1}

The focal source is \emph{SAGA: A Fast Incremental Gradient Method With Support for Non-Strongly Convex Composite Objectives} \citep{defazio2014saga}. Two records illustrate complementary uses of L1: a method description supports family membership, while a theoretical result supports condition-aware selection.

\klcard{klblue}{Source sentences | SAGA}{
\texttt{SAGA:19}\quad ``We start with some known initial vector $x^{0}\in\mathbb{R}^d$ and known derivatives $f_i'(\phi_i^0)\in\mathbb{R}^d$ with $\phi_i^0=x^0$ for each $i$.''\par\smallskip
\texttt{SAGA:20}\quad ``These derivatives are stored in a table data-structure of length $n$, or alternatively a $n\times d$ matrix.''\par\smallskip
\texttt{SAGA:58}\quad ``By using an unbiased update in SAGA, we are able to obtain a simple and tight theory, with better constants than SAG, as well as theoretical rates for the use of proximal operators.''\par\smallskip
\texttt{SAGA:29}\quad The displayed bound for the average iterate is
\[
\mathbb{E}[F(\bar{x}^{k})]-F(x^*)\leq\frac{4n}{k}
\left[\frac{2L}{n}\lVert x^0-x^*\rVert^2+f(x^0)
-\langle f'(x^*),x^0-x^*\rangle-f(x^*)\right].
\]
Here $\bar{x}^k=k^{-1}\sum_{t=1}^{k}x^t$ averages the iterates excluding $x^0$; $n$ is the number of finite-sum components, $k$ the iteration count, and $x^*$ an optimum. Each component $f_i$ is convex with $L$-Lipschitz continuous gradient; $F=f+h$ includes a convex regularizer $h$ with a computable proximal operator. The source specifies a step size of $1/(3L)$.
}

\klcard{klblue}{Scientific Fact | \texttt{saga.method}}{
\texttt{type:}\quad method description\par
\texttt{subject:}\quad SAGA\par
\texttt{statement:}\quad Maintains a table of previously computed component gradients and uses an unbiased corrected update.\par
\texttt{context:}\quad Incremental optimization; proximal support for composite objectives.\par
\texttt{evidence:}\quad Initialization: \texttt{SAGA:19}; gradient table: \texttt{SAGA:20}; unbiasedness and proximal support: \texttt{SAGA:58}. The update equation is at \texttt{SAGA:59} (not reproduced here).\par
\texttt{record note:}\quad Abridged method-identity record; the cited sentences are resolved through its source card and sentence map. Verification annotations are omitted from this card.\par
\texttt{reuse:}\quad Basis of SAGA's variance-reduction assignment; also cited by the family synthesis.
}

\klcard{klblue}{Scientific Fact | \texttt{saga.convex-rate}}{
\texttt{type:}\quad theoretical result\par
\texttt{subject:}\quad SAGA\par
\texttt{statement:}\quad The expected objective gap of the average iterate has the sublinear bound displayed above.\par
\texttt{context:}\quad Finite-sum structure; each component is convex and $L$-smooth; convex regularization with a computable proximal operator.\par
\texttt{evidence:}\quad Bound and step size: \texttt{SAGA:29}; finite-sum, convexity, and smoothness assumptions: \texttt{SAGA:9}; composite objective: \texttt{SAGA:10}.\par
\texttt{stored verification:}\quad Faithful extraction; source reference checked.\par
\texttt{reuse:}\quad Basis of the sublinear-rate assignment; qualifies the synthesis by a specific result and setting.
}

The theorem card uses the structured theoretical-result record. Its conditions stay attached to the bound when the fact is selected or compared. The method card and theorem card therefore contribute different kinds of evidence, even though both concern SAGA. A stored extraction verdict concerns fidelity to the cited sentences, not independent validation of the theorem. The saved records also contain a separate prose summary of this result, whose adjudication ended disputed; the structured record shown here was judged independently as faithful, and the two verdicts are retained side by side rather than reconciled.

\subsection{L2: From Facts to Domain Organization}
\label{app:l2}

The following cards and table illustrate four components of L2: a taxonomy, a canonical entity for SAGA, category assignments supported by its method description and theoretical results, and an evaluation relation to MNIST. The assignments retain their supporting facts, so labels derived from different results are not interpreted as a single combined guarantee.

\klcard{klgreen}{Domain Taxonomy | \texttt{variance-reduction}}{
\texttt{dimension:}\quad method family\par
\texttt{category:}\quad variance reduction\par
\texttt{criterion:}\quad A reference-point full gradient, per-example gradient table, or recursive gradient-difference correction is central to the method.\par
\texttt{positive examples:}\quad SAGA, SVRG, SARAH, SPIDER, SAG\par
\texttt{other dimensions:}\quad Assumption regime (e.g., convex, strongly convex); guarantee type (e.g., linear rate, sublinear rate).
}

\noindent
\begin{minipage}[t]{.485\linewidth}
\klcard{klgreen}{Canonical Entity | \texttt{SAGA}}{
\texttt{type:}\quad method\par
\texttt{name:}\quad SAGA\par
\texttt{alias:}\quad SAGA\par
\texttt{source:}\quad arXiv 1407.0202\par
\texttt{mention:}\quad Paper's method-name field.\par\smallskip
This record anchors one method identity; it does not demonstrate a multi-alias merge.
}
\end{minipage}\hfill
\begin{minipage}[t]{.485\linewidth}
\klcard{klgreen}{Category Assignment | \texttt{saga.family}}{
\texttt{subject:}\quad SAGA\par
\texttt{category:}\quad variance reduction\par
\texttt{basis:}\quad \texttt{saga.method}\par
\texttt{inference:}\quad system judgment\par
\texttt{certainty:}\quad confirmed\par
\texttt{validity:}\quad current\par
\texttt{reason:}\quad The gradient-table correction meets the family criterion.
}
\end{minipage}

\begin{table}[htbp]
\centering\small
\caption{Distinct organizational views of the same method. The assumption and guarantee rows share the convex result as their displayed basis. These labels describe that result, not every guarantee available for SAGA. The family assignment records three agreeing votes; all three displayed assignments have stored confirmed certainty, current validity, and passed evidence verification.}
\label{tab:app-saga-assignments}
\renewcommand{\arraystretch}{1.25}
\begin{tabular}{@{}>{\raggedright\arraybackslash}p{.23\linewidth}>{\raggedright\arraybackslash}p{.25\linewidth}>{\raggedright\arraybackslash}p{.46\linewidth}@{}}
\toprule
Dimension & Assigned category & Basis \\
\midrule
Method family & Variance reduction & \texttt{saga.method}: historical gradient-table correction \\
Assumption regime & Convex & \texttt{saga.convex-rate}: component assumptions at \texttt{SAGA:9} \\
Guarantee type & Sublinear rate & \texttt{saga.convex-rate}: convex result at \texttt{SAGA:29} \\
\bottomrule
\end{tabular}
\end{table}

\klcard{klgreen}{Typed Relation | SAGA paper $\rightarrow$ MNIST}{
\texttt{relation:}\quad evaluated on\par
\texttt{source / target:}\quad SAGA paper / MNIST dataset\par
\texttt{payload:}\quad Method: SAGA; task: binary classification.\par
\texttt{evidence locator:}\quad First dataset entry in the saved experiment card; source sentences \texttt{SAGA:162,166} identify MNIST among the evaluated datasets.\par
\texttt{interpretation:}\quad Records that the paper evaluates SAGA on this dataset and task; no comparative score is asserted.
}

Family selection makes SAGA relevant to a variance-reduction request. Comparing convergence guarantees additionally requires the L1 assumptions and measured quantities. Dataset relations provide another access route without implying that results on different datasets are directly comparable.

\subsection{L3: From Related Facts to a Scoped Synthesis}
\label{app:l3}

The final card demonstrates how a cross-paper account retains its coverage and evidence roles. It is an editorial revision of a stored family overview, narrowed to the evidence below; it is not a newly registered or automatically verified synthesis.

\klcard{klpurple}{Scoped Synthesis | \texttt{variance-reduction.overview}}{
\texttt{kind:}\quad method-family overview\par
\texttt{account:}\quad In the selected finite-sum studies, Prox-SVRG uses periodic full-gradient computations and SAGA maintains historical component gradients. Related work extends SVRG and SAGA to convex-concave saddle-point problems. Guarantees must be read with their assumptions, and uniform sampling may be inferior to an accelerated batch method in the reported experiments.\par\smallskip
\texttt{scope:}\quad The SAGA paper~\citep{defazio2014saga}; \emph{A Proximal Stochastic Gradient Method with Progressive Variance Reduction}~\citep{xiao2014proxsvrg}; and \emph{Stochastic Variance Reduction Methods for Saddle-Point Problems}~\citep{palaniappan2016saddle}. Only the mechanisms, results, and comparisons cited below are covered.\par\smallskip
\texttt{basis:}\quad Supporting and limiting evidence in Table~\ref{tab:app-synthesis-basis}; no opposing entry is asserted in this illustration.\par
\texttt{status:}\quad Editorially revised illustration. The original stored overview has review state \emph{unreviewed} and lifecycle state \emph{current}.
}

\begin{table}[htbp]
\centering\small
\caption{Evidence roles in the displayed synthesis. A supporting entry contributes to the account; a limiting entry qualifies its application. Locators identify sentences in each paper's saved source map. The convex-rate entry uses the corrected basis at \texttt{SAGA:29}.}
\label{tab:app-synthesis-basis}
\renewcommand{\arraystretch}{1.15}
\begin{tabular}{@{}>{\raggedright\arraybackslash}p{.15\linewidth}>{\raggedright\arraybackslash}p{.29\linewidth}>{\raggedright\arraybackslash}p{.50\linewidth}@{}}
\toprule
Role & Source fact & Contribution and source locator \\
\midrule
Supporting & Prox-SVRG method~\citep{xiao2014proxsvrg} & Periodic full-gradient computation; \texttt{Prox-SVRG:83} states when the full gradient is computed. \\
Supporting & \texttt{saga.method}~\citep{defazio2014saga} & Historical gradient storage; \texttt{SAGA:20}. Same fact as the L2 family-assignment basis. \\
Supporting & Saddle-point method~\citep{palaniappan2016saddle} & Extension to saddle-point problems: \texttt{Saddle:19}; convex-concave setting: \texttt{Saddle:14}. \\
Limiting & \texttt{saga.convex-rate}~\citep{defazio2014saga} & Specifies the guarantee in the convex setting (\texttt{SAGA:9,29}); this upper bound does not establish that faster convergence is impossible. \\
Limiting & Saddle-point comparison~\citep{palaniappan2016saddle} & Uniform-sampling SAGA/SVRG may be inferior to an accelerated batch method; \texttt{Saddle:118}. \\
\bottomrule
\end{tabular}
\end{table}


\endgroup

\section{Additional Details: Literature Retrieval}

\subsection{Fixed-Corpus Retrieval}
\label{app:fixed-corpus-retrieval}

\paragraph{Corpus and knowledge assets.}
The retrieval corpus comprises 64,183 papers, represented by a title--abstract index and six full-text Parquet shards~\citep{ajith2024litsearch}. Knowledge assets are constructed from a fixed, query-independent set of 900 source papers. The Full-text and ScholarStack conditions use the same source-paper set, which is not selected or adapted using query-specific gold papers. Membership in this set does not affect eligibility for retrieval: all conditions can search the entire corpus and inspect previously disclosed candidates under the same corpus-access limits.

\paragraph{Source-paper selection.}
Papers are eligible for asset construction if they have a nonempty title, a nonempty abstract, and a positive full-text character count. These criteria retain 55,631 of the 64,183 papers. The eligibility filter applies only to asset construction; the retrieval corpus retains all 64,183 papers. The source set consists of 720 semantic-coverage representatives (80\%) and 180 citation anchors (20\%).

Each eligible paper is represented by its title followed by a newline and its abstract. We construct two representations: (i) a 128-dimensional TF--IDF/SVD vector~\citep{salton1988term,deerwester1990indexing} and (ii) a 384-dimensional embedding produced by \texttt{all-MiniLM-L6-v2}~\citep{reimers2019sentence}. TF--IDF uses lowercasing, Unicode accent stripping, English stopword removal, word unigrams and bigrams, a minimum document frequency of 2, a maximum document frequency of 0.95, at most 50,000 features, and sublinear term frequency. Truncated SVD uses seven iterations with a fixed random seed. The two representations are independently normalized, scaled by $\sqrt{0.5}$, and concatenated into a 512-dimensional vector. We apply MiniBatchKMeans with 720 clusters, a batch size of 4,096, three initializations, at most 100 iterations, and a reassignment ratio of 0.01. 

Citation anchors are drawn from eligible papers outside the 720 representatives. They are ranked by descending incoming citation count within the corpus. The coverage and citation lists are interleaved so that the cumulative citation quota at position $r$ follows Python's $\mathrm{round}(0.2r)$, skipping identifiers already selected.

\paragraph{Deterministic baselines.}
BM25 uses SQLite FTS5 over titles and abstracts, with field weights of 4:1, at most 40 query tokens, and a top-100 output. Dense uses \texttt{all-MiniLM-L6-v2} with a maximum input length of 256 tokens and 384-dimensional embeddings. Paper inputs consist of the title, a newline, and the abstract, with missing fields replaced by empty strings. Query inputs use the original public query text without the BM25 token filter. Both paper and query embeddings are L2-normalized, allowing exact cosine search through dot products. Dense returns the top 100 papers. Papers with missing titles or abstracts remain in the corpus. Hybrid combines the two top-100 lists from BM25 and Dense, using equal-weight reciprocal-rank fusion with a rank constant of 60.

\paragraph{Additional details for ScholarStack.}
ScholarStack exposes existing record content without generating additional facts during retrieval. The agent receives a shortened textual rendering of the selected knowledge records for search planning. This rendering omits verification metadata; final ranking is grounded in independently inspected abstracts and full texts. L2 content includes assignment rationales and record titles. L3 content includes synthesis statements, scopes, and supporting explanations, without additional titles. Other records contribute substantive text while excluding non-informative fields such as identifiers, taxonomy bookkeeping, and serialized provenance. L2 or L3 records without substantive content produce empty text rather than metadata-only fallbacks. The selected content supports query formulation and evidence checking while preserving the distinct evidential roles of L1--L3.

An in-memory SQLite FTS5 index supports English-word and Chinese-character-bigram retrieval. The English and Chinese searches each return up to 60 records, which are merged using reciprocal-rank fusion with a rank constant of 60. Let $T(q)$ and $T(v)$ denote the query and record term sets, respectively, let $N$ be the number of indexed records, and let $\mathrm{df}(t)$ be the number of records containing term $t$. The primary ranking score is
\begin{equation}
S(q,v)
=
\sum_{t \in T(q) \cap T(v)}
\log \frac{N+1}{\mathrm{df}(t)+1}.
\label{eq:litsearch-content-score}
\end{equation}
Ties are resolved first by fused rank and then by a stable record identifier. Records must match at least $\min(2,|T(q)|)$ content terms; L3 records always require at least two. No layer is allocated a reserved result slot. Duplicate texts are removed after normalization, and at most two records are retained per source-paper combination.

For long records, the returned content is a 1,200-character window selected to maximize weighted query-term coverage. Candidate windows are drawn from positions near term matches, including the beginning of the text and up to the first eight occurrences of each query term. Each lookup returns at most three records. This procedure differs from corpus full-text reads, which return a window near the first matching term.

\paragraph{Agentic search protocols.}
Each query is processed through four model calls: planning, expansion, verification, and ranking. All Qwen3.8-Max runs disable thinking and use a temperature of 0. Within each model, prompts, tools, budgets, and final-context filtering are aligned across the compared conditions.

Table~\ref{tab:litsearch-stages} summarizes the per-stage tool allocations, and Table~\ref{tab:litsearch-budgets} reports the cumulative execution limits. Knowledge access can identify source candidates through bounded metadata resolution or motivate additional corpus searches. During verification, the agent distinguishes query conditions that are supported, unchecked, or contradicted. It prioritizes methodological details, comparisons, datasets, and results likely to affect the ranking, using one to three distinctive terms in each full-text request.

Only candidates disclosed before a model call can be inspected during that call; newly retrieved candidates become available for inspection in a subsequent stage. For final ranking, the model receives candidate metadata and direct evidence from abstracts or full texts. Intermediate knowledge records, discovery annotations, and the preceding gap checklist are excluded. The model returns at most 40 unique corpus identifiers encountered during retrieval. Direct support takes precedence over unchecked details, and missing evidence is not treated as contradiction. The same final-context policy applies to all aligned experimental conditions.

\begin{table}[t]
\centering
\small
\caption{Maximum numbers of tool calls per stage. Asset lookups are available to Full-text and ScholarStack but not to the no-asset control. All allocations are subject to the cumulative limits in Table~\ref{tab:litsearch-budgets}.}
\label{tab:litsearch-stages}
\begin{tabular}{lrrrrr}
\toprule
Stage
& \shortstack{Corpus\\searches}
& \shortstack{Title\\lookups}
& \shortstack{Asset\\lookups}
& \shortstack{Abstract\\inspections}
& \shortstack{Full-text\\reads} \\
\midrule
0: Initial planning     & 3 & 2 & 1 & 0 & 0 \\
1: Expansion and checks & 3 & 2 & 1 & 6 & 4 \\
2: Final verification   & 2 & 2 & 0 & 4 & 2 \\
3: Ranking              & 0 & 0 & 0 & 0 & 0 \\
\bottomrule
\end{tabular}
\end{table}

\begin{table}[t]
\centering
\small
\caption{Execution and response limits per query. Byte limits are measured in UTF-8; character and token limits are enforced separately.}
\label{tab:litsearch-budgets}
\begin{tabular}{@{}p{0.43\linewidth}p{\dimexpr0.57\linewidth-2\tabcolsep\relax}@{}}
\toprule
Resource & Limit \\
\midrule
Model calls / candidate-context size
& 4 / 180 papers \\
Corpus searches / title lookups
& 8 (20 papers per search) / 6 \\
Metadata resolutions / abstract inspections
& 10 / 10 \\
Asset lookups and returned content
& 2 lookups; at most 3 records per lookup; 8,000 characters in total \\
Full-text reads
& 6 calls covering at most 6 distinct papers \\
Full-text response length
& 4,000 characters per call; 6,000 per paper; 24,000 in total \\
Model-output size
& 18,000 bytes per call \\
Prompt size
& 160,000 bytes per call; 480,000 in total \\
Corpus-response length
& 160,000 characters in total \\
Observed input / output token stopping thresholds
& 140,000 / 16,000 tokens \\
Wall-clock time
& 300 seconds per model call; 720 seconds per query \\
Automatic retries / format repairs
& 0 / 0 \\
\bottomrule
\end{tabular}
\end{table}

\paragraph{Metrics.}
Runtime and total input-plus-output token usage are averaged over queries and exclude knowledge construction. Dense runtime excludes offline encoding and loading. Hybrid runtime is the sum of the component-query and fusion times. Qwen token-usage records are complete, whereas the supplementary GPT token counts are lower bounds based on reported usage.

Let $Q$ denote the evaluation query set, $G_q$ the gold-paper set for query $q$, $P_q^{(k)}$ the first $k$ predictions, and $r_q$ the rank of the first gold-paper match. We compute
\begin{align}
\mathrm{R@}k
&=
\frac{1}{|Q|}
\sum_{q \in Q}
\frac{|G_q \cap P_q^{(k)}|}{|G_q|},
\\
\mathrm{MRR@40}
&=
\frac{1}{|Q|}
\sum_{q \in Q}
\begin{cases}
1/r_q, & \text{if a gold paper is ranked within the top 40},\\
0, & \text{otherwise}.
\end{cases}
\end{align}
Recall is the macro-average of gold-paper coverage, not the fraction of queries with at least one hit. Predictions are finalized before scoring, and evaluation does not use an LLM-based relevance judge.

\paragraph{Additional results.}
We report additional results using another base LLM. Table~\ref{tab:retrieval-gpt} reports fixed-corpus results with GPT-5.6 Terra. Table~\ref{tab:retrieval-coverage-gpt} further reports results by knowledge-source coverage.

\begin{table}[t]
\centering
\caption{Fixed-corpus retrieval results with GPT-5.6 Terra. Asset size denotes the number of papers used to construct the auxiliary assets. Runtime and token usage are averaged per query; GPT token counts are reported-usage lower bounds. Bold indicates the best value in each retrieval-metric column.}
\label{tab:retrieval-gpt}
\begin{tabular}{lrrrrrrr}
\toprule
Method
& \shortstack{Corpus}
& \shortstack{Asset Size}
& R@1 & R@5 & R@10 & MRR@40
& Tokens \\
\midrule
Full-text
& 64,183 & 900
& 0.670 & 0.773 & 0.773 & 0.721 & 41.1k \\
ScholarStack
& 64,183 & 900
& \textbf{0.710} & \textbf{0.803} & \textbf{0.803} & \textbf{0.763} & 39.5k \\
\bottomrule
\end{tabular}%
\end{table}

\begin{table}[t]
\centering
\caption{Fixed-corpus retrieval results by knowledge-source coverage with GPT-5.6 Terra. Bold indicates the best value in each column, including ties.}
\label{tab:retrieval-coverage-gpt}
\setlength{\tabcolsep}{4pt}
\begin{tabular}{@{}lrrrrrrrr@{}}
\toprule
& \multicolumn{4}{c}{In-source}
& \multicolumn{4}{c}{Out-of-source} \\
\cmidrule(lr){2-5}
\cmidrule(lr){6-9}
Method
& R@1 & R@5 & R@10 & MRR@40
& R@1 & R@5 & R@10 & MRR@40 \\
\midrule
Codex raw (w/o full-text)
& 0.605 & 0.719 & 0.719 & 0.684
& 0.677 & 0.774 & 0.774 & 0.720 \\
Full-text
& \textbf{0.711} & 0.772 & 0.772 & 0.763
& 0.645 & 0.774 & 0.774 & 0.695 \\
ScholarStack
& \textbf{0.711} & \textbf{0.798} & \textbf{0.798} & \textbf{0.781}
& \textbf{0.710} & \textbf{0.807} & \textbf{0.807} & \textbf{0.753} \\
\bottomrule
\end{tabular}
\end{table}

\subsection{Open-World Retrieval}
\label{app:open-corpus-retrieval}

\paragraph{Evaluation setting and domain routing.}
We evaluate open-world retrieval on a fixed subset of 89 queries from the open-ended SAGE benchmark. The subset is sampled using fixed seeds and comprises 29 computer-science, 28 natural-science, and 32 healthcare queries. Each query expresses a research-oriented information need rather than a paper title.

During evaluation, the benchmark-provided domain label routes each query to its corresponding frozen domain collection. This deterministic routing removes domain-classification errors as a confound in the comparison. Retrieval prompts and parameters are not specialized by domain. Gold titles and relevance labels are withheld until the returned lists are finalized.

\paragraph{Query-independent source collections.}
Each domain collection is constructed independently of SAGE queries and labels. We first define 18 broad topic groups per domain to cover major subfields rather than benchmark-specific entities. Generic explicit and implicit topic queries retrieve candidate papers from an internal scholarly index that combines multiple retrieval algorithms. We reserve a coverage quota of 45 papers per topic and rank the remaining candidates using retrieval provenance, topic coverage, bibliographic metadata, citation evidence, and full-text availability.

Candidate full texts must contain at least 2,000 characters and ten parsed sentences. Deterministic quality checks exclude a small number of malformed documents, yielding the source-paper counts in Table~\ref{tab:open-corpus-assets}. Within each domain, Full-text and ScholarStack use exactly the same source-paper identifiers.

\begin{table}[t]
\centering
\caption{Frozen source collections for open-world retrieval. "Source papers" denotes the number of full-text documents underlying both the Full-text and ScholarStack conditions.}
\label{tab:open-corpus-assets}
\begin{tabular}{lrr}
\toprule
Domain & SAGE queries & Source papers \\
\midrule
Computer science & 29 & 1,754 \\
Natural science  & 28 & 1,749 \\
Healthcare       & 32 & 1,748 \\
\bottomrule
\end{tabular}
\end{table}

\paragraph{Knowledge-layer construction.}
Each domain collection is processed once with Qwen3.8-Max (thinking disabled) to construct L1--L3. L1 extracts paper-grounded statements, conditions, methods, findings, limitations, and sources. L2 organizes L1 records into categories, assignments, relations, and citation links. L3 builds scoped cross-paper syntheses from compatible L1 evidence, retaining links to supporting or limiting evidence. Invalid or weakly grounded objects are not promoted.

The resulting stores and hashes are frozen before retrieval scoring and reused across queries, so construction cost is excluded from per-query retrieval time.

\paragraph{Source-matched Full-text baseline.}
Full-text first searches titles and abstracts, returning at most ten papers within a 3,000-character budget. The agent then selects one to five papers and reads their full texts, returning at most two passages per paper within 9,000 characters. Full-text therefore uses one catalog call and one read call, with at most 12,000 characters of local context.

\paragraph{ScholarStack access.}
ScholarStack uses the same paper-first interface and budgets. Its first call searches a compiled structured catalog and returns at most ten candidate papers within 3,000 characters. After selecting one to five identifiers, the second call returns up to 9,000 characters of query-focused L1 facts, grounded L2 structure, and provenance-linked L3 syntheses. Thus, both conditions expose different representations of the same source-paper collection under the same 12,000-character ceiling.

Local knowledge is accessed before external retrieval and provides aliases, conditions, relationships, comparison axes, and citation anchors for search and verification. It changes search planning and evidence allocation, but not the final candidate set.

\paragraph{Shared agentic-search framework.}
After the local catalog-and-read phase, both conditions use the same bounded multi-round search agent. The agent decomposes the request, runs keyword and exact-title searches, expands candidates through citations, and consults public web evidence to resolve aliases or gaps. The output is a tiered set of papers, with Tier~1 containing about five highest-confidence recommendations.

Both conditions use Qwen3.8-Max with thinking disabled, the Balanced retrieval policy, live network retrieval, disabled query caches, and the same soft ten-minute time budget.

\paragraph{Title matching and head metrics.}
Evaluation uses the Tier~1 set explicitly produced by the agent rather than the first five entries of a flattened result list. Let $\mathcal{Q}$ denote the query set, $H_q$ the Tier~1 predictions for query $q$, and $G_q$ its gold-paper set.

Titles are normalized by lowercasing and removing non-alphanumeric characters. Matching requires equality between normalized titles, with a containment exception for long subtitle variants: the shorter normalized title must be contained in the longer title and exceed 80\% of its length. Let $C_q \subseteq G_q$ denote the unique gold papers matched by $H_q$. Per-query head recall, precision, and F1 are
\begin{align}
R_q &= \frac{|C_q|}{|G_q|},
&
P_q &= \frac{|C_q|}{|H_q|},
&
F_q &= \frac{2P_qR_q}{P_q+R_q}.
\end{align}
A metric is defined as zero when its denominator is zero. Reported values are query-level macro-averages:
\begin{equation}
\operatorname{Macro}(X)
=
\frac{1}{|\mathcal{Q}|}
\sum_{q \in \mathcal{Q}} X_q,
\qquad X \in \{R,P,F\}.
\end{equation}
Macro F1 is therefore the mean of the per-query F1 scores, not the harmonic mean of macro precision and macro recall.

\paragraph{Relevance-weighted recall.}
SAGE distinguishes two \emph{most relevant} seed papers from additional \emph{relevant} shared references. We assign each gold paper the weight
\begin{equation}
w_q(g)
=
\begin{cases}
2, & \text{if } g \text{ is most relevant},\\
1, & \text{if } g \text{ is relevant}.
\end{cases}
\end{equation}
The relevance-weighted head-recall metrics are
\begin{align}
\operatorname{MacroWR}
&=
\frac{1}{|\mathcal{Q}|}
\sum_{q \in \mathcal{Q}}
\frac{\sum_{g \in C_q} w_q(g)}
     {\sum_{g \in G_q} w_q(g)},
\\
\operatorname{MicroWR}
&=
\frac{\sum_{q \in \mathcal{Q}} \sum_{g \in C_q} w_q(g)}
     {\sum_{q \in \mathcal{Q}} \sum_{g \in G_q} w_q(g)}.
\end{align}

\subsection{Details: Novelty Assessment}
\label{app:novelty}

\paragraph{System and judge.}
Both methods use the same base model (Qwen3.8-Max, thinking disabled) and agentic executor with a fixed tool budget. Web search is disabled, so the comparison is closed-book. The two methods differ only in knowledge access: \textbf{Full-text} applies a general, domain-agnostic novelty method while reading the original related-paper texts; \textbf{ScholarStack} applies the same method over assets compiled offline from those texts. Ground-truth matching uses GPT-6 with high reasoning effort and no Web access.

\paragraph{Benchmark composition.}
The benchmark is built from a citation-linked scientific corpus and spans 12 non-CS-dominated fields. It comprises $85$ cases: $40$ \emph{overlap-present} targets, each a paper with genuine prior-art overlaps, and $45$ \emph{absent} targets that test whether a system correctly reports no genuine overlap. Table~\ref{tab:novelty-domains} gives the per-field breakdown of the present targets. Each present target provides a research contribution (its problem, method family, and setting) and a candidate pool of related papers (mean $14.4$ per case: on average $3.4$ genuine prior works plus $\approx\!11$ same-domain non-cited \emph{hard distractors}); absent targets carry a comparable pool (mean $11.0$) with no genuine overlap. Across the $40$ present targets there are $135$ gold overlaps (mean $3.4$, range $2$--$6$ per target), each recorded as a (cited work, overlap axis, residual difference) tuple. By axis, $106$ are core---same problem ($44$), similar method family ($35$), similar mechanism ($27$)---and $29$ are peripheral---shared data or setting ($14$), other ($15$); this distribution motivates the $1.0/0.3$ core/peripheral weighting used below.

\begin{table}[htbp]
\centering\small
\caption{Per-field breakdown of the $40$ overlap-present novelty targets (12 fields).}
\label{tab:novelty-domains}
\begin{tabular}{lc@{\qquad}lc@{\qquad}lc}
\toprule
Field & Count & Field & Count & Field & Count \\
\midrule
Linguistics & 5 & Physics    & 4 & Education        & 2 \\
Medicine    & 5 & Psychology & 4 & Art              & 1 \\
Engineering & 5 & Business   & 4 & Chemistry        & 1 \\
Mathematics & 4 & Biology    & 4 & Materials Sci. & 1 \\
\bottomrule
\end{tabular}
\end{table}

\paragraph{Metric provenance.}
The weighted identification F1 matches each citation-derived gold overlap to the agent output with the LLM matcher. Let each overlap carry axis weight $w{=}1.0$ for core axes (same problem, similar method family, similar mechanism), which determine whether a contribution is novel, and $w{=}0.3$ for peripheral axes (shared data/setting, other), which are incidental to novelty; a uniform scheme would score an incidental shared-dataset overlap as heavily as a same-mechanism one. The matcher labels each gold overlap hit ($1.0$), partial ($0.5$), or miss ($0.0$); a proposed overlap matching no gold is a false positive. Per case, weighted recall is $\sum_{\text{gold}} w\cdot\mathrm{credit} / \sum_{\text{gold}} w$ and weighted precision is $\sum_{\text{matched proposed}} w\cdot\mathrm{credit} / \sum_{\text{proposed}} w$ (credit $\in\{1,0.5\}$); F1 is their harmonic mean, and a zero denominator (no gold, or no proposed overlap) yields $0$ for that component. Reported values are macro-averaged over the 40 overlap-present targets. Gold overlaps are the target's cited prior work; citation is an operational proxy for genuine prior art. Because gold is citation-derived, our conclusions are relative comparisons under identical gold labels.

\section{Additional Details: Scientific Question Answering}

\subsection{Single-paper QA}
\label{app:single-qa}

This appendix reports benchmark composition and evaluation details for single-paper question answering, followed by the full set of evaluated conditions, a per-domain breakdown of the results, and qualitative examples.

\paragraph{Benchmarks and evaluation.}
We follow the evaluation metrics of the original QASPER~\citep{dasigi-etal-2021-dataset} and PeerQA~\citep{baumgartner-etal-2025-peerqa} benchmarks. For QASPER (416 papers / 1{,}428 questions), we use the official evaluator to compute token-level Answer-F1, overall and by answer type, and paragraph-level Evidence-F1, taking the best reference score per question before averaging. For PeerQA (208 papers / 579 questions), we report the original answerability metrics---per-class F1, accuracy, weighted F1, and macro F1---on 495 labeled questions, and generation ROUGE-L against free-form answers (FF) and annotated evidence (AE) on the 245-question generation subset. We additionally report mean query-time input tokens as a cost measure, excluding knowledge construction. Human denotes the reported human-answer reference on QASPER; on PeerQA, it denotes model answers using human-annotated evidence, with answerability evaluated only on answerable questions.

\paragraph{Complete condition set (PeerQA).}
Table~\ref{tab:peerqa-allcond} reports the evaluated conditions using the metric-specific subsets above. RAG@$k$ places the top-$k$ retrieved passages in context; the main text reports RAG@10, while RAG@20 doubles that passage budget. ScholarStack improves the aggregate answerability metrics over RAG@10 at lower input usage, with a small decrease in AE ROUGE-L. RAG@20 achieves higher Macro-F1 (0.6273 versus 0.6134) and generation scores than ScholarStack, using approximately 2.6 times as many input tokens.

\begin{table}[htbp]
\centering
\small
\caption{All evaluated conditions on PeerQA (208 papers / 579 questions). RAG@$k$ retrieves the top-$k$ passages. Answerability is evaluated on 495 questions: Ans-F1 and Unans-F1 denote F1 for answerable and unanswerable questions; Acc., W-F1, and Macro-F1 denote accuracy, class-frequency-weighted F1, and unweighted mean class F1. Generation ROUGE-L is evaluated on 245 questions against free-form reference answers (FF) and annotated evidence (AE). Human denotes model answers using human-annotated evidence, with answerability scored only on answerable questions. Tokens denotes mean input tokens per question; \textbf{bold} marks ScholarStack's improvements over RAG@10.}
\label{tab:peerqa-allcond}
\resizebox{\linewidth}{!}{%
\begin{tabular}{lccccccc c}
\toprule
& \multicolumn{5}{c}{Answerability} & \multicolumn{2}{c}{Generation (Rouge-L)} & \\
\cmidrule(lr){2-6}\cmidrule(lr){7-8}
Method & Ans-F1 & Unans-F1 & Acc. & W-F1 & Macro-F1 & FF & AE & Tokens \\
\midrule
Full-text                & 0.806 & 0.507 & 0.721 & 0.738 & 0.656 & 0.155 & 0.205 & 12{,}577 \\
Human \emph{(reference)} & 0.732 & --    & --    & --    & --    & 0.150 & 0.161 & 301 \\
\midrule
RAG@10                  & 0.736 & 0.476 & 0.649 & 0.677 & 0.606 & 0.142 & 0.156 & 1{,}864 \\
RAG@20                  & 0.766 & 0.489 & 0.679 & 0.703 & 0.627 & 0.149 & 0.184 & 3{,}481 \\
ScholarStack             & \textbf{0.749} & \textbf{0.478} & \textbf{0.661} & \textbf{0.687} & \textbf{0.613} & \textbf{0.145} & 0.156 & \textbf{1{,}350} \\
\bottomrule
\end{tabular}%
}
\end{table}

\paragraph{Per-domain robustness (PeerQA).}
Table~\ref{tab:peerqa-domain} reports the full answerability panel and answer-generation Rouge-L for each PeerQA domain. On the two domains with substantial support, machine learning and NLP, which together account for 446 of the 495 answerability questions (about 90\%), ScholarStack is competitive or stronger: it slightly exceeds RAG on machine learning (0.594 versus 0.591 Macro-F1) and surpasses both RAG and the full-text reader on NLP (0.725 versus 0.690 and 0.707). The geoscience and biomedicine-and-social-science splits are small (27 and 22 answerability questions, with only 1 and 10 unanswerable, and 14 and 4 generation questions respectively), so their per-domain scores carry high variance; within this limitation, ScholarStack again leads on the biomedicine-and-social-science split and trails on geoscience. The advantage of ScholarStack therefore holds on the domains with sufficient support and is not driven by any single field.

\begin{table}[htbp]
\centering
\small
\caption{Per-domain results on PeerQA. Parentheses give the numbers of answerability / generation questions. Ans-F1 and Unans-F1 denote F1 for answerable and unanswerable questions; Acc., W-F1, and Macro-F1 denote accuracy, class-frequency-weighted F1, and unweighted mean class F1. RgL$*{\mathrm{FF}}$ and RgL$*{\mathrm{AE}}$ denote generation ROUGE-L against free-form reference answers and annotated evidence, respectively. \textbf{Bold} marks the best Macro-F1 in each domain.}
\label{tab:peerqa-domain}
\small
\begin{tabular}{lccccccc}
\toprule
Method & Ans-F1 & Unans-F1 & Acc. & W-F1 & Macro-F1 & RgL$*{\mathrm{FF}}$ & RgL$*{\mathrm{AE}}$ \\
\midrule
\multicolumn{8}{l}{\emph{Machine learning} (313 answerability / 158 generation)} \\
Full-text  & 0.808 & 0.517 & 0.725 & 0.747 & \textbf{0.662} & 0.142 & 0.199 \\
RAG        & 0.722 & 0.460 & 0.633 & 0.666 & 0.591 & 0.130 & 0.149 \\
ScholarStack  & 0.729 & 0.459 & 0.639 & 0.672 & 0.594 & 0.128 & 0.141 \\
\midrule
\multicolumn{8}{l}{\emph{NLP} (133 / 69)} \\
Full-text  & 0.822 & 0.593 & 0.752 & 0.761 & 0.707 & 0.194 & 0.235 \\
RAG        & 0.777 & 0.604 & 0.714 & 0.731 & 0.690 & 0.174 & 0.181 \\
ScholarStack  & 0.822 & 0.628 & 0.759 & 0.771 & \textbf{0.725} & 0.193 & 0.212 \\
\midrule
\multicolumn{8}{l}{\emph{Geoscience} (27 / 14)} \\
Full-text  & 0.818 & 0.200 & 0.704 & 0.795 & \textbf{0.509} & 0.105 & 0.131 \\
RAG        & 0.791 & 0.182 & 0.667 & 0.768 & 0.486 & 0.112 & 0.124 \\
ScholarStack  & 0.700 & 0.143 & 0.556 & 0.679 & 0.421 & 0.099 & 0.066 \\
\midrule
\multicolumn{8}{l}{\emph{Biomedicine \& social science} (22 / 4)} \\
Full-text  & 0.667 & 0.000 & 0.500 & 0.364 & 0.333 & 0.150 & 0.185 \\
RAG        & 0.625 & 0.000 & 0.455 & 0.341 & 0.313 & 0.130 & 0.149 \\
ScholarStack  & 0.645 & 0.154 & 0.500 & 0.422 & \textbf{0.400} & 0.134 & 0.094 \\
\bottomrule
\end{tabular}%
\end{table}

\paragraph{Qualitative examples.}
Table~\ref{tab:qa-examples} presents selected qualitative examples with abridged reference answers and model responses. ScholarStack and Full-text provide relevant answers in these cases, although their completeness differs, while RAG gives a different answer or abstains. 

\begin{table}[htbp]
\centering
\small
\caption{Selected qualitative examples on QASPER and PeerQA. Gold denotes the reference answer. Reference answers and model responses are abridged for presentation; answer completeness varies. Aggregate quality and input usage are reported in Tables~\ref{tab:qasper-main} and~\ref{tab:peerqa-main}.}
\label{tab:qa-examples}
\resizebox{\linewidth}{!}{%
\begin{tabular}{p{0.24\linewidth} p{0.20\linewidth} p{0.20\linewidth} p{0.16\linewidth} p{0.20\linewidth}}
\toprule
Question & Gold & Full-text & RAG & ScholarStack \\
\midrule
\multicolumn{5}{l}{\emph{QASPER}} \\
What inference models are used? & XGBoost, AdaBoost, Random Forest & XGBoost, AdaBoost and Random Forest & Gaussian Processes & XGBoost, AdaBoost, Random Forest \\
How many actions are present in the dataset? & 14{,}769 & 14{,}769 & No Answer & 14{,}769 \\
\midrule
\multicolumn{5}{l}{\emph{PeerQA}} \\
Does the optimization happen only over map locations observed so far? & No; it also covers unobserved (hallucinated) regions & No; any region observed or hallucinated & No Answer & No; goals from any observed or hallucinated region \\
What values are represented in Table 2? & Zero-shot classification performance across 5 datasets and 14 language models & Zero-shot classification performance for all models and datasets & No Answer & Performance statistics for all models and datasets \\
\bottomrule
\end{tabular}%
}
\end{table}

\subsection{Multi-Paper QA}
\label{app:multi-paper-qa}

\paragraph{Ablation design.}
MDAQA requires evidence synthesis across scientific papers. Each question provides gold answers, which we used to identify the corresponding gold papers and construct the Full-text and knowledge-layer inputs. The benchmark provides English questions and answers, with pre-extracted paper content available through the associated SPIQA materials. We evaluate the same 797 questions used in the main comparison. Table~\ref{tab:mdaqa-ablation-cost} reports the full MDAQA ablation. The comparisons below examine three knowledge configurations using the same answering model, Qwen3.8-Max with reasoning disabled, and the same ordinary prompt. The configurations share a frozen base evidence package for each question, produced by the same retrieval and filtering process, with no additional retrieval during answer generation. L1 provides selected source-grounded facts. L1+L2 adds domain categories and assignments, and the full ScholarStack configuration adds persistent cross-paper syntheses through L3. Full-text serves as the reference method based on parsed papers.

\paragraph{Judging and provenance audit.}
GPT-6 Astra with high reasoning effort scores each candidate independently using the question and a fixed evidence package. Gold answers (not shown in the table) enter evaluation only as additional anonymous candidates and are not supplied as answer-generation inputs or treated as ground truth. Candidate identifiers and presentation order are randomized, with no pairwise ranking or scoring by similarity to the gold answers. Content scores are frozen before provenance auditing. The audit checks key claims against the evidence actually supplied to each configuration, distinguishing full support, partial support, no support, contradiction, and common knowledge. The existence of reference identifiers is verified programmatically, while semantic support is assessed separately. Invalid or missing executions are recorded as N/A, and empty or refused answers remain part of the results. Each question is treated as an independent statistical unit.

\begin{table}[htbp]
\centering
\small
\caption{Full ablation and generation-cost results on the 797-question MDAQA bank. Bare denotes the answering model using the ordinary prompt and supplied evidence, without the additional skill instructions. Skill denotes Cross-Paper Evidence Synthesis skill, which guides comparison of study conditions, synthesis of findings, and attribution of claims to evidence. Corr., Comp., Synth., Cal., and Dir. denote correctness, completeness, cross-paper synthesis, calibration, and directness. Scores are weighted means on a 0--4 scale, except Total (0--20). Generation-cost columns use the available per-question usage audit. Construction and judging are excluded. Bold marks the best value in each column, where higher is better for the six score columns and lower is better for the two cost columns.}

\label{tab:mdaqa-ablation-cost}
\resizebox{\linewidth}{!}{%
\begin{tabular}{lrrrrrrrr}
\toprule
Method & Corr. & Comp. & Synth. & Cal. & Dir. & Total & Tokens & Sec. \\
\hline
Full-text + Bare & 1.11 & 2.01 & 1.41 & 0.92 & 3.01 & 8.46 & 20.8k & 13.13 \\
ScholarStack (L1) + Bare & 3.90 & 3.56 & 2.99 & 3.78 & 3.82 & 18.04 & \textbf{4.1k} & 9.83 \\
ScholarStack (L1) + Skill & 3.91 & 3.60 & 3.03 & 3.82 & 3.83 & 18.18 & 4.2k & 10.45 \\
ScholarStack (L1+L2) + Bare & 3.82 & 3.25 & 2.80 & 3.26 & \textbf{3.88} & 17.01 & 4.8k & \textbf{7.20} \\
ScholarStack (L1+L2) + Skill & 3.89 & 3.53 & 3.02 & 3.78 & 3.84 & 18.05 & 4.9k & 10.67 \\
ScholarStack (L1+L2+L3) + Bare & 3.89 & 3.35 & 3.48 & 3.90 & 3.80 & 18.41 & 5.5k & 7.54 \\
ScholarStack (L1+L2+L3) + Skill & \textbf{3.92} & \textbf{3.62} & \textbf{3.59} & \textbf{3.96} & 3.76 & \textbf{18.85} & 5.7k & 10.51 \\
\bottomrule
\end{tabular}%
}
\end{table}

\paragraph{Effects of knowledge layers and answering procedure.}
L1, L2, and L3 differ in how much comparison and interpretation they provide. L1 makes selected findings and study conditions more accessible; L2 groups related findings but still requires the model to judge comparability; L3 adds explicit scope, evidence, and limitations. Adding L2 reduces output from 557.12 to 378.98 tokens, but completeness and calibration decline.

With the same evidence, Cross-Paper Evidence Synthesis (CPES) improves answer scores. The largest Total gain occurs with L1 + L2, from 17.01 to 18.05. L3 already supplies some comparisons and qualifications, so its additional gain is smaller. Overall, CPES helps organize evidence, preserve qualifications, and improve answer quality.

\section{Additional Details: Evidence-Grounded Generation}
\subsection{Related-Work Generation}
\label{app:related-work}

\subsubsection{Benchmark and Controlled Conditions}

We sample 64 eligible targets from the flattened-sections OARelatedWork test split, each with complete outputs under all executed generation conditions. The final cohort contains 47 CS/ML papers and 17 papers from other domains. Eligibility requires nonempty target text, at least five distinct cited papers, valid nonduplicated reference identifiers, and non-abstract full text for every reference. The complete-case selection uses generation completion only; it does not inspect generated quality, judge scores, or the held-out related-work section.

The benchmark supplies a fixed cited-paper collection for each target, and the author-written related-work section is excluded from generation. Both conditions use independent Qwen3.8-Max writer sessions with reasoning disabled. Full-text and ScholarStack otherwise share the same target material, citation registry, task instructions, writing procedure, and 300--600-word constraint. Full-text exposes the original reference files. ScholarStack instead exposes frozen paper-level assets, L1 source-grounded facts, L2 categories and relations, and L3 scoped syntheses, and cannot reread or silently substitute the reference files. The comparison therefore changes the evidence representation while holding the writer and generation procedure fixed.

\subsubsection{Evaluation Protocol}

\paragraph{Independent quality scoring.}
An LLM judge evaluates the 64 sections generated by each of the two methods, submitting each section in a separate request. Each request contains one anonymized draft, the focal-manuscript context, the original reference materials, and the citation map; it does not contain the competing draft, the generation condition, the ScholarStack assets, or the held-out author section. The judge assigns integer scores from 1 to 5 for five dimensions. Relevance measures alignment with the focal manuscript. Coverage measures whether important related work and information are included. Specificity measures concrete descriptions of methods, contributions, and differences. Support measures whether the supplied references substantiate the generated statements. Overall is a holistic judgment constrained by the component scores. Structured outputs are validated against the frozen schema, including resolution of cited source and draft spans.

\paragraph{Factuality and citation overlap.}
The factuality audit checks generated claims against the original reference texts and assigns True, Wrong Citation, Unverifiable, or False. Label proportions are computed within each generated section before being averaged across the 64 targets, so a section containing more claims does not receive greater weight. Citation F1 compares the generated citation set with the references cited by the held-out author section. It measures agreement with the author-selected reference set, not whether every generated statement is supported and not whether the set exhausts all relevant work.
\subsection{Literature Review Generation}
\label{app:review-protocol}

\subsubsection{Resource Accounting}
\label{app:review-accounting}

We report the mean resources used to produce one review, including literature retrieval and survey generation. Input and output are shown in thousands of tokens, and cache hit is the fraction of input served from cache. The cost column applies a common token-price calculation to the recorded usage so that the two workflows can be compared on the same reporting surface.

\begin{table}[htbp]
\centering
\small
\caption{Per-report resource use and standardized cost for the two locally evaluated review-generation workflows. k denotes one thousand tokens.}
\label{tab:review-resource-cost}
\setlength{\tabcolsep}{6pt}
\begin{tabular}{lrrrr}
\toprule
System & \shortstack{Input(k)} & \shortstack{Output(k)} & \shortstack{Cache hit(\%)} & \shortstack{Cost(USD)} \\
\midrule
Claude Code DR & 3,128 & 106 & 78.94 & 7.18 \\
\textbf{ScholarStack} & 10,533 & 117 & 96.02 & 3.49 \\
\bottomrule
\end{tabular}
\end{table}

ScholarStack processes a broader evidence set and therefore uses more input tokens than Claude Code DeepResearch (10,533k versus 3,128k), while their output usage is similar (117k versus 106k). Its substantially higher cache-hit rate, 96.02\% versus 78.94\%, reduces the price contribution of repeated context. Under the standardized calculation, this yields 3.49 USD per report for ScholarStack and 7.18 USD for Claude Code DeepResearch. These values exclude reusable knowledge construction and evaluation, and should be interpreted as normalized estimates rather than actual invoices or end-to-end deployment costs.

\subsubsection{Trace-Backed Excerpt Comparisons}
\label{app:review-cases}
We randomly sample one paired example for each of L1, L2, and L3. Each example presents excerpts from the two reports on the same task in stacked panels. The panels retain the reports' wording, with headings, line breaks, and emphasis added for readability; numeric citations are the reports' original identifiers. We verified all quoted statistics against the source PDFs.

\definecolor{reviewours}{HTML}{F3EEE8}

\paragraph{L1: preserving conditions and limits of technical claims.}
Both reports discuss Space2Vec using the same source paper. Claude Code states the broad efficacy claim and later notes that scale hyperparameters require tuning. ScholarStack places the direction and magnitude of the wavelength effect, a negative setting, and two applicability boundaries next to the claim.

\begin{figure}[H]
\centering
\setlength{\fboxsep}{5pt}
\noindent\fcolorbox{black!25}{black!2}{\begin{minipage}{0.94\linewidth}\small
\textbf{Claude Code DR}\par\vspace{2pt}\hrule\vspace{4pt}
Empirically, the multi-scale Space2Vec representation is reported to outperform RBF kernels, multi-layer feed-forward nets, and tile embedding approaches on POI type prediction and geo-aware image classification, precisely because those baselines handle only a single scale while Space2Vec handles all scales [10].\par\smallskip
\textbf{Later limitation}\quad The number and spacing of scales are hyperparameters requiring tuning.
\end{minipage}}\par\vspace{6pt}
\noindent\fcolorbox{black!25}{reviewours}{\begin{minipage}{0.94\linewidth}\small
\textbf{ScholarStack}\par\vspace{2pt}\hrule\vspace{4pt}
The improvement is, however, conditional on the wavelength hyperparameter: relative MRR gains over the wrap baseline are $\textbf{-0.2\%}$, $\textbf{+0.6\%}$ and $\textbf{+2.1\%}$ for $\lambda_{\min}$ of 1 km, 500 m and 50 m respectively, so one setting is a small regression and the benefit grows as the minimum wavelength shrinks [1].\par\smallskip
Space2Vec's location encoding becomes less important when rich context from nearby points is already available. The model is formulated for 2-D continuous spaces, with higher-dimensional extension unexplored, and only late fusion with image classifiers is investigated [1].
\end{minipage}}
\caption{L1 example on Space2Vec. ScholarStack keeps effect size, hyperparameter setting, negative result, and applicability boundaries adjacent to the claim.}
\label{fig:review-l1}
\end{figure}

\paragraph{L2: organizing the literature across multiple dimensions.}
Both reports organize graph-based time-series anomaly detection along graph construction and detection objective. Claude Code states the two axes directly, whereas ScholarStack explains why a single axis would merge scientifically different methods and keeps architecture as a cross-cutting dimension.

\begin{figure}[H]
\centering
\setlength{\fboxsep}{5pt}
\noindent\fcolorbox{black!25}{black!2}{\begin{minipage}{0.94\linewidth}\small
\textbf{Claude Code DR}\par\vspace{2pt}\hrule\vspace{4pt}
\textbf{Axis A --- How the graph is obtained.} Prior/explicit graphs are derived from known topology or physical connectivity; learned graphs are inferred from data via attention, similarity embeddings, or top-$k$ neighbor selection; dual/hybrid graphs jointly capture inter-variable and intra-variable (temporal) structure.\par\smallskip
\textbf{Axis B --- The training/scoring paradigm.} Reconstruction-based methods learn to reconstruct normal windows; anomalies yield high reconstruction error. In adversarial/generative methods, a generator models the normal distribution and discriminator/reconstruction residuals score anomalies. Prediction-based methods forecast the next value from graph context; large forecast error flags anomalies. Self-supervised or contrastive methods use pretext tasks or contrastive objectives to learn normal representations.
\end{minipage}}\par\vspace{6pt}
\noindent\fcolorbox{black!25}{reviewours}{\begin{minipage}{0.94\linewidth}\small
\textbf{ScholarStack}\par\vspace{2pt}\hrule\vspace{4pt}
A single-dimension taxonomy would be misleading: the same paradigm appears with static similarity graphs, learned implicit graphs, and dynamic graphs, and the same graph strategy appears with forecasting, reconstruction, and flow objectives.\par\smallskip
This survey therefore uses \textbf{graph construction as the primary axis} (what the graph is and where it comes from) and \textbf{detection objective as the secondary axis} (how a score is produced from it), with architecture treated as a cross-cutting implementation choice.
\end{minipage}}
\caption{L2 example on graph-based time-series anomaly detection. ScholarStack explains why multiple dimensions are needed and how they determine the survey structure.}
\label{fig:review-l2}
\end{figure}

\paragraph{L3: tracing how later methods address earlier limitations.}
Both reports summarize prompt-based continual learning as an evolution from selecting prompts to composing them differentiably. ScholarStack extends this into a cross-paper lineage: each successor is tied to the limitation it addresses, and the sequence is summarized as a shift in where adaptation occurs.

\begin{figure}[H]
\centering
\setlength{\fboxsep}{5pt}
\noindent\fcolorbox{black!25}{black!2}{\begin{minipage}{0.94\linewidth}\small
\textbf{Claude Code DR}\par\vspace{2pt}\hrule\vspace{4pt}
Prompt-based methods add a small, learnable, input-conditioned instruction mechanism, buying plasticity while keeping the backbone frozen. The internal evolution from \textbf{L2P's hard pool selection} to \textbf{DualPrompt's complementary decomposition} to \textbf{CODA-Prompt's end-to-end differentiable assembly} is essentially a progressive removal of non-differentiable bottlenecks.\par\smallskip
Hard selection is simpler but caps plasticity, while differentiable assembly improves plasticity at the cost of complexity.
\end{minipage}}\par\vspace{6pt}
\noindent\fcolorbox{black!25}{reviewours}{\begin{minipage}{0.94\linewidth}\small
\textbf{ScholarStack}\par\vspace{2pt}\hrule\vspace{4pt}
\textbf{CODA-Prompt} improves over DualPrompt by attacking the non-differentiable key-query selection that limits plasticity, and over L2P by replacing pool selection with attention-weighted decomposition.\par\smallskip
\textbf{HiDe-Prompt} improves over L2P by attacking the degradation of prompt methods under self-supervised pre-training. \textbf{LAE} treats prompting as one PET instantiation and adds calibration, accumulation, and ensembling. \textbf{EASE} makes the subspace per task and synthesizes old-class prototypes inside new subspaces.\par\smallskip
The direction of travel is consistent: each step moves the locus of adaptation further away from the backbone weights, first into prompts, then into prompts plus statistics, then into per-task modules, and finally into the classifier and inference-time aggregation rule.
\end{minipage}}
\caption{L3 example on continual learning with pre-trained models. ScholarStack connects successive methods through the limitations they address and extracts a coherent direction of travel.}
\label{fig:review-l3}
\end{figure}


\subsection{Experimental-Design Generation}
\label{app:expdesign}

\paragraph{System and judge.}
All methods use the same base model (Qwen3.8-Max, non-thinking), a fixed tool budget, and a closed-book setting, differing only in knowledge access: Base LLM reads raw files with no task method, Full-text adds a domain-agnostic design method over the same raw files, and ScholarStack applies that method over assets compiled offline from the related corpus. Coverage matching and the blind pairwise judge use GPT-6 (high reasoning effort, no web access).

\paragraph{Benchmark composition.}
The benchmark is built from a citation-linked scientific corpus and spans 12 non-CS-dominated fields, with $n{=}38$ targets. Table~\ref{tab:expdesign-domains} gives the per-field breakdown. Each target is a paper that reports an experimental study; it provides a research contribution to validate and a candidate pool of related papers (median $14$ per case, range $13$--$17$) that a system may draw on when designing experiments. Ground truth is the set of experiments the target paper actually performed, extracted from its source text---$220$ gold experiments in total (mean $5.8$, median $4$, range $1$--$20$ per target), each a (name, purpose, kind) tuple. By kind, $93$ are core validation experiments (main comparisons $54$, ablations $39$), $119$ are supporting studies (analysis $82$, robustness $17$, theory $15$, human evaluation $5$), and $8$ are other; this distribution motivates the kind weighting defined below.

\begin{table}[htbp]
\centering\small
\caption{Per-field breakdown of the $38$ experimental-design targets (12 fields).}
\label{tab:expdesign-domains}
\begin{tabular}{lc@{\qquad}lc@{\qquad}lc}
\toprule
Field & Count & Field & Count & Field & Count \\
\midrule
Medicine    & 5 & Physics    & 4 & Education        & 2 \\
Engineering & 5 & Business   & 4 & Art              & 1 \\
Linguistics & 4 & Biology    & 4 & Chemistry        & 1 \\
Mathematics & 4 & Psychology & 3 & Materials Sci. & 1 \\
\bottomrule
\end{tabular}
\end{table}

\paragraph{Metric provenance.}
The gold experiments are the experiments the target paper actually performed, extracted from the source paper by GPT-6; they are not human-validated and are not an exhaustive enumeration of all valid designs, so absolute coverage is bounded by design under-determination and the reported effects should be read as relative differences between methods under identical conditions. Coverage F1 assigns each gold experiment kind a weight $w$ by how directly it tests a contribution's central claims: main comparisons and ablations ($1.0$), supporting analysis, robustness, human-evaluation, and theory studies ($0.6$), and other ($0.3$). The judge labels each gold experiment hit ($1.0$), partial ($0.5$), or miss ($0.0$); per case, weighted recall is $\sum_{\text{gold}} w\cdot\mathrm{credit} / \sum_{\text{gold}} w$ and precision is the fraction of proposed experiments that match some gold experiment (a partial match counts as matched; experiments matching no gold are ``extras''), with F1 the harmonic mean of precision and weighted recall and a zero denominator yielding $0$ for that component; reported values macro-average per-case F1/recall/precision over the $38$ targets. The blind pairwise judge receives the contribution and two anonymized designs in a per-case-seeded random order, with an explicit ``more is not better; ignore formatting'' instruction; wins/losses/ties are decoded after judging (Table~\ref{tab:expdesign-pref}).

\paragraph{Precision, not padding.}
Base LLM's low precision ($0.255$) reflects over-proposal: it emits a mean of $22.9$ experiments per target, versus $18.6$ (Full-text) and $18.7$ (ScholarStack). The blind judge, instructed that more is not better, prefers the shorter claim-targeted designs, which is why both methods beat Base LLM by wide margins ($32/38$, $34/38$) despite producing fewer experiments.

\paragraph{Limitations.}
Limitations include moderate sample ($n{=}38$), a single judge family used for both coverage matching and pairwise preference, self-constructed benchmarks, and GPT-6-extracted (non-human-validated) gold whose design under-determination is discussed above. The coverage-F1 margin between Full-text and ScholarStack is small ($0.006$) and we report no significance tests, so we do not read it as a quality difference; the effects we rest on are the wide pairwise preference of both methods over Base LLM and ScholarStack's lower per-query cost.

\subsection{Idea Generation}
\label{app:idea-generation-evaluation}

\subsubsection{Benchmark composition}
\begin{table}[H]
\centering
\caption{Domain composition of the idea-generation benchmark.}
\label{tab:idea-generation-domains}
\small
\begin{tabular}{lr}
\toprule
Domain & Count \\
\midrule
AI and engineering & 14 \\
Chemistry and materials & 12 \\
Life sciences and medicine & 4 \\
Public health and social sciences & 3 \\
Environmental science & 2 \\
\midrule
Total & 35 \\
\bottomrule
\end{tabular}
\end{table}

\subsubsection{Evaluation dimensions}
\begin{table}[H]
\centering
\caption{Definitions of evaluation dimensions from \citet{li2024chainofideas}.}
\label{tab:idea-generation-criteria}
\small
\begin{tabular}{p{0.17\linewidth} p{0.76\linewidth}}
\toprule
Dimension & Definition \\
\midrule
Novelty & Are the problems or approaches new? Is this a novel combination of familiar techniques? Is it clear how this work differs from previous contributions? Is related work adequately referenced? \\
Significance & Is the idea important? Are other people (practitioners or researchers) likely to use these ideas or build on them? Does the idea address a difficult problem in a better way than previous research? Does it provide a unique theoretical or pragmatic approach? \\
Clarity & Is the paper clearly written? Is it well-organized? Does it adequately inform the reader? \\
Feasibility & Can the idea be realized with existing technology or methods? Are there any technical difficulties or bottlenecks? Is the idea clear and logical? Are there any obvious errors or unreasonable parts in the idea, and can the experiments be designed normally according to this idea? \\
Effectiveness & How likely is the proposed idea going to work well (e.g., better than existing baselines)? \\
\bottomrule
\end{tabular}
\end{table}

\subsubsection{Score aggregation}

For each case, the LLM judge receives the task, common reference papers, and three anonymized outputs. It ranks complete outputs independently on each dimension, with instructions not to favor response length or the number of proposed ideas.
The ranking and aggregation procedure is specific to this evaluation. For case $i$, configuration $g$, and dimension $d$, let $r_{i,g,d}\in\{1,2,3\}$ denote the rank assigned by the judge. The corresponding score, mean dimension score, and overall mean are
\begin{equation}
s_{i,g,d}=\frac{3-r_{i,g,d}}{2},\qquad
S_{g,d}=\frac{1}{N}\sum_{i=1}^{N}s_{i,g,d},\qquad
\overline{S}_{g}=\frac{1}{5}\sum_{d=1}^{5}S_{g,d},
\end{equation}
where $N=35$. Thus, first, second, and third place receive 1, 0.5, and 0, respectively, with equal weight assigned to each case and dimension. The resulting scores express relative preference among the evaluated configurations.

\section{Additional Details: Claim and Consistency Assessment}

\subsection{Scientific Claim Verification}
\label{app:verifier-claim}

\paragraph{Dataset and relation labels.}
The $80$ claim--paper pairs come from the public training/development split of NLPCC 2026 Shared Task 10, Track 2~\citep{nlp2ct2026task10}, with $20$ instances per label. Claims and papers are deduplicated, and source texts are limited to $50{,}000$ characters. This is a selected public subset, not the official held-out test. \emph{Supported} requires direct support from the cited paper; \emph{Overstate} exceeds the scope or strength of that support; \emph{Topical Match} shares the subject matter without supporting the specific claim; and \emph{Irrelevant} lacks a substantive relation. The labels describe support by the cited source, not scientific truth.

\paragraph{Matched paper contexts.}
Full-text uses the complete article text supplied by the dataset, subject to its existing extraction quality. ScholarStack reuses the original paper-level L1 objects unchanged, without supplemental extraction, appended source passages, or claim-conditioned selection. The entire collection is supplied directly; no query-time retrieval or L2/L3 processing is performed. Both conditions use the same source papers and receive neither task images nor external retrieval.

\paragraph{Model and scoring.}
Both conditions use Qwen3.8-Max with thinking disabled and the same classification prompt, without access to gold labels. Predicted labels are scored by exact match, without an LLM judge. Accuracy is the number of correct predictions divided by $80$, with failures and abstentions retained in the denominator. All $80$ instances in each condition have a valid final prediction.

\paragraph{Cost accounting.}
For condition $c$, mean query usage is
\begin{equation}
\overline{T}_{c}=\frac{1}{80}\sum_{a\in\mathcal{A}_{c}}
\left(T^{\mathrm{in}}_{a}+T^{\mathrm{out}}_{a}\right),
\end{equation}
where $\mathcal{A}_{c}$ includes all evaluation attempts, including retries. Total usage is $963{,}703$ tokens for the Full-text and $225{,}368$ for ScholarStack, giving the means in Table~\ref{tab:verifier-claim} and a relative reduction of $1-225{,}368/963{,}703=76.61\%$. Historical construction usage for these objects is unavailable. The reported savings therefore concern queries against existing assets and do not establish end-to-end savings or a construction-cost break-even point.

\paragraph{Scope and limitations.}
The evaluation covers relation classification, not evidence localization or the official joint classification-and-evidence score. The original objects lack a complete field-level evidence map, so correct labels do not establish source traceability or rationale faithfulness. Extraction may omit qualifications needed to distinguish support from overstatement or topical similarity. The balanced, length-limited public subset does not represent the original label distribution or longer papers, and the single-run accuracy difference has not been tested for significance. Evaluation on additional held-out papers is needed to assess reliability and evidence fidelity.

\section{Details of the Continuous Research Experiments}
\label{app:continuous}

This appendix supplements Section~\ref{sec:continuous-research} with research questions, shared task requirements, metric definitions, and a case on the \emph{Conversational QA} topic illustrating reuse within and across task chains.

\subsection{Topics and Questions}
\label{app:continuous-questions}

Five topics were drawn from the QA area of LitSearch. Topic labels (which papers belong to each topic) were reviewed manually and were available only to the evaluator. Each topic has an initial question and a distinct, dependent follow-up question (Table~\ref{tab:app-continuous-questions}). The follow-up introduces a more specific research objective for which earlier findings remain relevant. Both questions use the same five-stage protocol to compare reuse across conditions; this fixed sequence is an evaluation design, not a required workflow for ScholarStack. A ten-item checklist of expected findings is used only to evaluate reuse in the follow-up.

\begin{table}[htbp]
\centering\small
\caption{Initial and follow-up research questions for the five topics. Each follow-up question builds on findings from the initial chain of the same topic.}
\label{tab:app-continuous-questions}
\renewcommand{\arraystretch}{1.1}
\setlength{\tabcolsep}{3pt}
\begin{tabular}{@{}>{\raggedright\arraybackslash}p{0.17\linewidth}>{\raggedright\arraybackslash}p{0.38\linewidth}>{\raggedright\arraybackslash}p{0.40\linewidth}@{}}
\toprule
Topic & First-chain question & Follow-up question (abridged) \\
\midrule
Adversarial distractors and unanswerable questions & How can extractive reading-comprehension models be made robust to adversarial distractor sentences and to unanswerable questions at the same time, without sacrificing accuracy on answerable questions? & Propose one combined training-and-inference recipe (unanswerable augmentation, distractor-aware selection, calibrated abstention) and specify how to measure transfer between the two robustness settings. \\
\addlinespace[2pt]
Retriever and reader in open-domain QA & How should the retriever be trained or supervised so that the passages it returns are the ones the reader can actually use, rather than the ones that merely contain the answer string? & Design one retriever training objective that uses per-passage reader utility as the label, and specify its validation against hard-negative training under equal compute and index size. \\
\addlinespace[2pt]
Multi-hop QA & What retrieval and reasoning strategies let multi-hop QA systems find and compose evidence across documents without exploiting shortcut cues, and how can this be measured? & Design one evaluation-plus-training protocol that measures and reduces disconnected reasoning, specifying benchmark-construction controls and the retrieval route. \\
\addlinespace[2pt]
Conversational QA & How can conversational QA systems resolve dependencies on dialogue history (ellipsis, coreference, topic shift) so that answer quality does not degrade as the conversation gets longer? & Design one training-and-evaluation protocol under self-predicted history, decide with evidence which history-handling route to adopt, and specify how to measure degradation with conversation length. \\
\addlinespace[2pt]
Retrieval-augmented generation & How can a generator be trained or prompted to rely on retrieved passages when they are relevant and to ignore or flag them when they are irrelevant or conflicting, and how should this behaviour be evaluated? & Design one generator training-and-evaluation protocol that decides when to rely on retrieved passages under irrelevant or conflicting evidence, with a validated reliance metric. \\
\bottomrule
\end{tabular}
\end{table}

\subsection{Shared Task Requirements and Checks}
\label{app:continuous-instructions}

\paragraph{Common requirements.}
Each stage receives shared rules, a task brief, earlier deliverables, and the condition's tools; Notes and ScholarStack start fresh sessions. Outputs use tool-returned evidence, link numbers and conditions to source sentences, and distinguish reported statements, inferences, and hypotheses. Deliverables follow stage-specific JSON schemas. All conditions run a self-check for unsupported numbers, unresolved sentence references, and out-of-scope papers.

\paragraph{Asset registration.}
ScholarStack additionally checks source references and numerical grounding before registering task-derived objects. Statements without a source remain notes rather than facts. These checks support traceability but do not establish human review or semantic correctness.

\subsection{Metric Definitions}
\label{app:continuous-metrics}

\begin{itemize}
\item \textbf{Recall@100.} Number of papers in the topic's manually reviewed relevant-paper set that appear among the first 100 retrieval results, divided by the size of that relevant-paper set.
\item \textbf{Citation coverage.} Number of papers in the same relevant-paper set cited in the review's per-paper findings, divided by the size of that set.
\item \textbf{RACE.} Four dimensions (comprehensiveness, insight, instruction following, readability), each 0--10, scored by an LLM judge with thinking disabled; the three conditions are anonymized and shuffled per topic.
\item \textbf{Idea quality.} Mean of five dimensions (novelty, significance, clarity, feasibility, expected effectiveness), each 1--5, for the selected idea.
\item \textbf{Overlap support.} For each overlap claimed in novelty assessment, the judge reads the cited sentences and labels the overlap \emph{supported}, \emph{partial}, or \emph{unsupported}; the metric is the share labelled supported.
\item \textbf{Design overall.} A holistic score (0--4), assessed alongside six dimensions: problem and contribution clarity, evidence and condition consistency, novelty analysis, testability and claim coverage, fairness and discriminability, and feasibility and economy. The overall score is assigned separately rather than averaged from the dimension scores.
\item \textbf{Checklist compliance.} The number of passed binary checks divided by 17 for a complete chain, averaged across topics. The checklist covers literature review generation (4 items), idea generation (3), novelty assessment (3), and experimental-design generation (7). Checks assess recorded conditions, evidence references, stated hypotheses and assumptions, decision--output consistency, and the presence of evaluation controls and hypothesis--experiment links. They measure compliance with output requirements, not semantic correctness or overall scientific quality. For the experimental-design layer comparison, compliance is evaluated across the chain after replaying experimental-design generation with upstream outputs fixed.
\item \textbf{Correct reuse coverage.} Each follow-up question has ten expected paper--aspect findings. Coverage is the fraction cited for the specified aspect in any follow-up deliverable, counting each finding once, averaged over five topics in Table~\ref{tab:continuous-reuse}. ``Correct'' denotes paper--aspect matching. This measures output coverage; access logs and costs help distinguish reuse from independent recovery.
\item \textbf{Object accesses.} Queries log returned objects and their originating chain and stage. First-chain object accesses count objects originating in the initial chain and returned during the follow-up, including repeated returns; distinct-object counts deduplicate identifiers within each chain. The L3 read share is the fraction of these first-chain object accesses that return L3 syntheses. An access records an object returned by a query, not necessarily its use in the final output.
\item \textbf{Tokens and input change.} New tokens sum non-cached input and output over every model call in a chain. Per-stage non-cached input counts input tokens excluding cache hits for each of the five stages. Input change is the percentage change in mean total input tokens from the initial to the follow-up chain, including cached input in both.
\item \textbf{Raw-document reads.} Number of calls to the raw-paper reading tool, including repeated reads. Parenthetical paper counts give distinct papers read, and sentence counts sum returned sentences across calls. Queries for registered evidence sentences are not raw-document reads.
\item \textbf{Commands and knowledge-layer queries.} Commands count logged tool-command invocations across a chain; knowledge-layer queries count calls to the knowledge-layer interface. Both include repeated calls.
\end{itemize}

\begingroup
\raggedbottom
\setlength{\intextsep}{8pt}
\subsection{Construction Cost}
\label{app:continuous-cost}

The knowledge layer is compiled once for the paper collection and shared by every chain: 25.0M tokens in total (23.0M input, 2.0M output). Chain costs in Section~\ref{sec:continuous-research} include the write-back and checking performed during a chain. Table~\ref{tab:app-continuous-cost} totals the input and output of the two chains of all five topics for the Context condition and for ScholarStack with and without construction.

\begin{table}[htbp]
\centering\small
\caption{Total input and output tokens over the two chains of the five topics (millions). Cached input is included; construction is counted once.}
\label{tab:app-continuous-cost}
\begin{tabular}{@{}lrr@{}}
\toprule
 & Context & ScholarStack \\
\midrule
First chains (5) & 92.6 & 22.8 \\
Follow-up chains (5) & 297.2 & 20.4 \\
Ten chains & 389.8 & 43.2 \\
Ten chains plus construction & --- & 68.3 \\
\bottomrule
\end{tabular}
\end{table}

For a single topic, the first chain costs 18.5M for Context and 29.6M for ScholarStack including construction; after the follow-up chain the totals are 78.0M and 33.7M, so construction is recovered within the second chain. As Table~\ref{tab:app-continuous-cost} shows, Context's cost grows with every additional chain because the whole history is re-sent (92.6M for the five first chains, 297.2M for the five follow-up chains), whereas ScholarStack's follow-up chains cost no more than its first chains (20.4M against 22.8M).

We further examine construction token usage in a separate preliminary study with collections of 10--100 papers (Figure~\ref{fig:app-construction-scaling}). Across the tested sizes, total token usage grows approximately linearly, while average usage remains around 60k tokens per paper. These results suggest relatively stable per-paper construction overhead within the evaluated range.

\begin{figure}[!htbp]
\centering
\includegraphics[width=\linewidth]{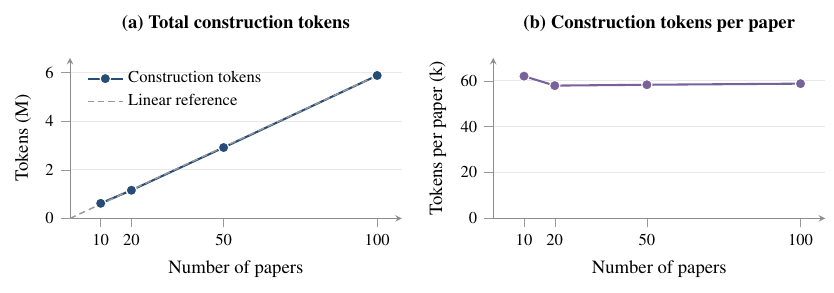}
\caption{Preliminary construction token usage for independent builds of the L1--L3 knowledge layer. (a) Total input and output tokens; the dashed line is a proportional reference through the largest observed point, not a fitted trend. (b) Average tokens per paper. Cached input is counted once.}
\label{fig:app-construction-scaling}
\end{figure}

\subsection{A Worked Example: \emph{Conversational QA}}
\label{app:continuous-example}

We use \emph{Conversational QA}, the fourth topic in Table~\ref{tab:app-continuous-questions}, as a worked example of reusing evidence-linked judgments and revising them when new evidence is found. Figure~\ref{fig:app-continuous-example} gives an overview. Table~\ref{tab:app-continuous-stages} describes the understanding used and retained in the initial chain; Table~\ref{tab:app-continuous-arms} summarizes access costs across its five stages.

\begin{figure}[!htbp]
\centering
\includegraphics[scale=1,trim=5pt 5pt 5pt 5pt,clip]{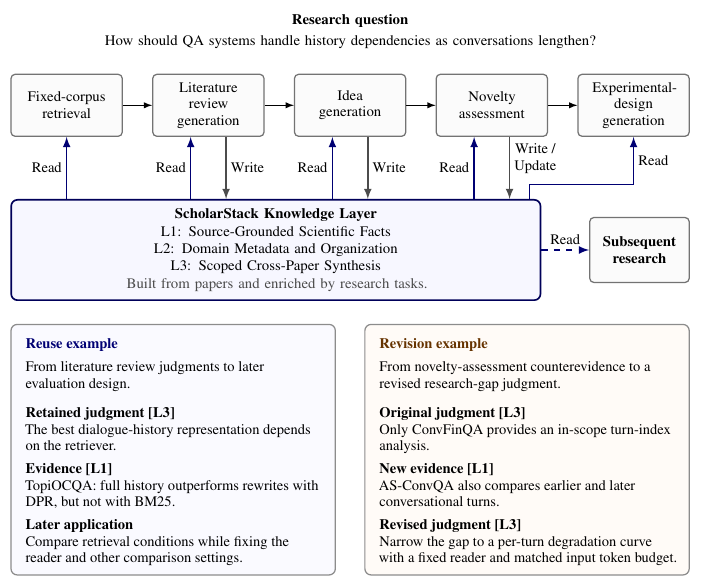}
\caption{Cross-task understanding reuse and revision on the \emph{Conversational QA} topic. Read arrows retrieve assets; Write arrows register new objects; Update revises earlier judgments. Idea-generation writes retain hypotheses, not established findings. Retrieval and experimental-design generation do not write back in this example. The dashed extension denotes subsequent research, evaluated with the same five-stage protocol. The reuse and revision cards summarize the examples in Figures~\ref{fig:app-continuous-design-reuse} and~\ref{fig:app-continuous-revision-trace}, respectively.}
\label{fig:app-continuous-example}
\end{figure}

\begin{table}[!htbp]
\centering\footnotesize
\caption{Understanding used and retained in the first chain of the \emph{Conversational QA} topic. Task outputs and registered assets remain distinct; proposed ideas are retained as hypotheses.}
\label{tab:app-continuous-stages}
\renewcommand{\arraystretch}{1.05}
\setlength{\tabcolsep}{3pt}
\begin{tabular}{@{}>{\raggedright\arraybackslash}p{0.22\linewidth}>{\raggedright\arraybackslash}p{0.33\linewidth}>{\raggedright\arraybackslash}p{0.40\linewidth}@{}}
\toprule
Task & Existing understanding used & Retained output or asset \\
\midrule
Fixed-corpus retrieval & Compiled research entities and evidence about relevant datasets and methods. & Ranked papers with research roles and sub-questions; no new assets registered. \\
Literature review generation & Retrieved findings, study conditions, and source sentences. & Supplementary facts, term aliases, and scoped syntheses, including the need to account for dialogue-history source and retriever when comparing results. \\
Idea generation & Review themes, open problems, and conflicting findings. & Candidate hypotheses, including a segment-scoped memory that gates information by provenance. \\
Novelty assessment & Evidence on topic segmentation, cross-turn memory, and history selection. & A narrower proposal for bounded memory conditioned on topic shifts, evidence-linked overlap judgments, revised syntheses, and additional candidate baselines. \\
Experimental-design generation & Evidence about datasets, baselines, metrics, and evaluation conditions. & A test plan with hypotheses, controls, and cited evaluation choices; no new assets registered. \\
\bottomrule
\end{tabular}
\end{table}

Table~\ref{tab:app-continuous-arms} shows that ScholarStack accesses retained knowledge and registered evidence without raw-document rereads in the initial chain. In the follow-up chain, which asks for a training-and-evaluation protocol under self-predicted history, it reuses the initial chain's comparison criterion concerning history source and retriever. Queries return first-chain objects 121 times; knowledge-layer queries fall from 144 to 70, with no raw-document reads and similar new-token use to the first chain. The examples below illustrate how retained judgments support later work and how new evidence leads to their revision.

\begin{table}[!htbp]
\centering\small
\caption{The three conditions on the \emph{Conversational QA} topic (first chain). Per-stage inputs are non-cached input tokens in thousands.}
\label{tab:app-continuous-arms}
\setlength{\tabcolsep}{4pt}
\renewcommand{\arraystretch}{1.1}
\begin{tabular*}{\linewidth}{@{\extracolsep{\fill}}lccc@{}}
\toprule
 & Context & Notes & ScholarStack \\
\midrule
Commands & 60 & 110 & 98 \\
Raw-document reads (papers; sentences) & 185 (58; 17,596) & 88 (43; 1,383) & 0 \\
Where understanding is kept & session history & 5 note files & knowledge layer \\
\midrule
\multicolumn{4}{@{}l}{\textit{Per-stage non-cached input (k)}} \\
Fixed-corpus retrieval & 60 & 50 & 53 \\
Literature review generation & 157 & 116 & 125 \\
Idea generation & 171 & 52 & 40 \\
Novelty assessment & 210 & 78 & 77 \\
Experimental-design generation & 245 & 75 & 65 \\
\bottomrule
\end{tabular*}
\end{table}

\paragraph{Evidence-grounded reuse.}
Figure~\ref{fig:app-continuous-design-reuse} illustrates how a retained judgment informs later work. During experimental-design generation, a task-specific Knowledge View retrieves retained judgments from the knowledge layer together with their evidence references. One review judgment states that the relative performance of full dialogue history and question rewriting depends on the retriever. Following its Evidence References locates the comparison in the TopiOCQA paper: full history performs better with DPR, whereas rewrites perform better with BM25. The final design varies BM25, a task-trained dense retriever, and oracle passages, while fixing the reader and other comparison conditions. This links a reused interpretation to source evidence and concrete experimental controls, supporting accuracy and traceability without treating registration as proof of correctness.

\begin{figure}[!htbp]
\centering
\includegraphics[width=\linewidth]{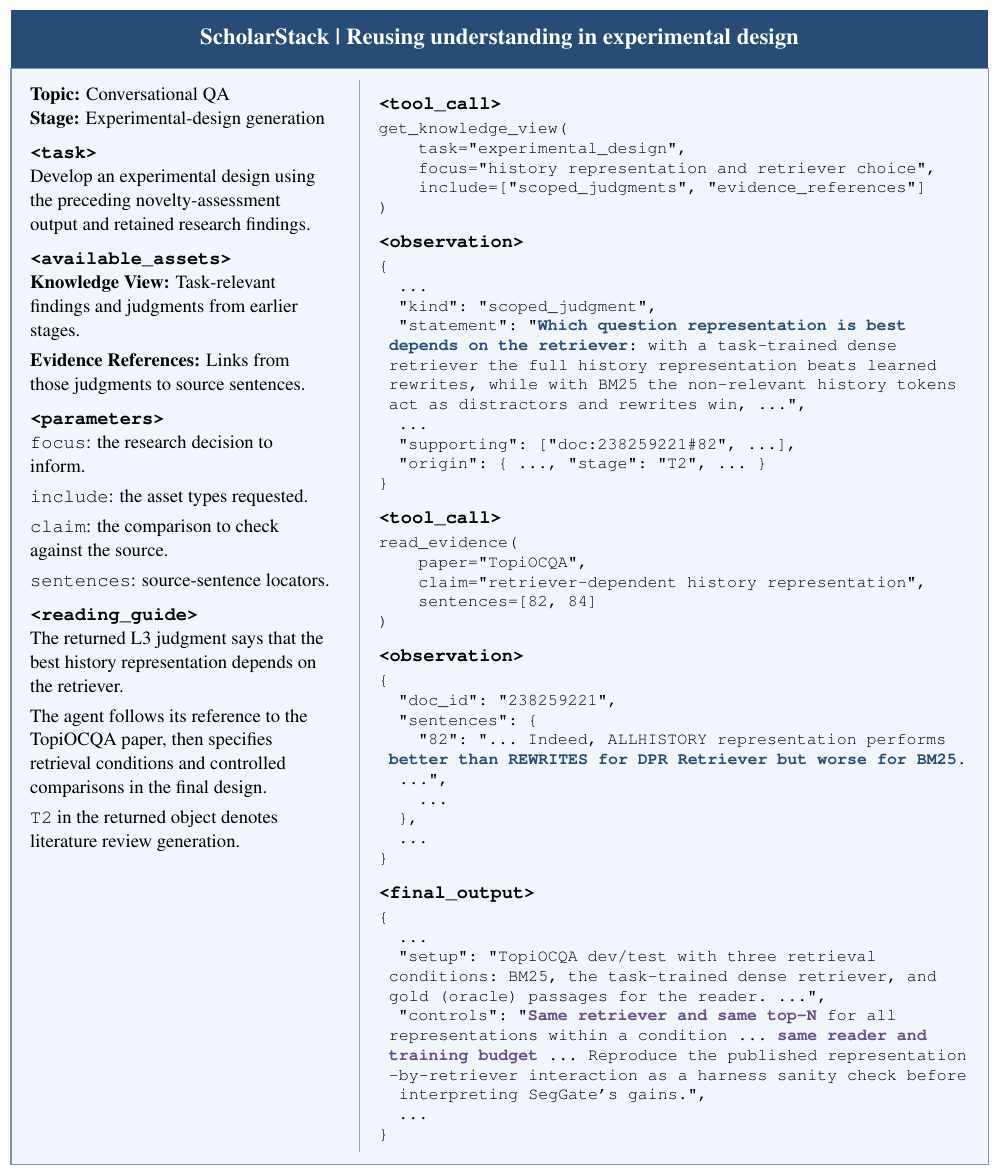}
\caption{An example of evidence-grounded reuse. The left column summarizes the task and parameters; the right shows a Knowledge View lookup, an evidence lookup, and the resulting design excerpt. JSON records are excerpts from the actual run. Ellipses mark omissions.}
\label{fig:app-continuous-design-reuse}
\end{figure}

\begin{samepage}
For the same experimental-design stage, the two baselines recover earlier findings through different routes. Context carries prior discussion in its conversation history and supplements it by reading a study of confidence- and uncertainty-based answer selection in conversational QA. Notes first reads saved novelty and idea notes and locates paper references in the review notes. It then searches for TopiOCQA evaluation details and reads multiple passages from that paper and other sources. Notes also specifies appropriate retriever and reader controls. ScholarStack accesses the retained judgment through a Knowledge View and checks its linked evidence directly.
\par
\end{samepage}

Table~\ref{tab:app-continuous-design-case} summarizes the design-stage comparison. ScholarStack uses fewer new tokens than either baseline, with an overall design score matching Notes and exceeding Context. Together with the trace, this illustrates reuse of an evidence-linked judgment at lower processing cost; the stage-level comparison does not isolate the effect of a single tool call.

\begin{table}[!htbp]
\centering
\small
\setlength{\tabcolsep}{4pt}
\renewcommand{\arraystretch}{1.15}
\caption{Design-stage comparison for the \emph{Conversational QA} case. Scores use the 0--4 experimental-design scale; bold marks best values. Raw-document reads exclude lookups of registered evidence sentences.}
\label{tab:app-continuous-design-case}
\begin{tabular}{@{}lp{4.6cm}rrr@{}}
\toprule
Method & Access to earlier findings & New tokens (k) & Score & Raw reads \\
\midrule
Context & Conversation history and supplementary source reading & 366.9 & 2 & 3 \\
Notes & Notes, paper search, and passage reading & 101.8 & \textbf{3} & 14 \\
ScholarStack & Knowledge View and linked evidence sentences & \textbf{88.5} & \textbf{3} & \textbf{0} \\
\bottomrule
\end{tabular}
\end{table}

\clearpage
\paragraph{Evidence-driven revision.}
During novelty assessment, new evidence leads to revisions of two judgments formed during an earlier review. Figure~\ref{fig:app-continuous-revision-trace} presents the original registration payloads, source response, and recorded revision event for one such correction, with the corresponding source-lookup and write-back commands. The revised judgment narrows the proposed gap to per-turn degradation under a fixed reader and matched input budget, providing an evidence-linked premise for subsequent evaluation design.

\begin{figure}[!htbp]
\centering
\includegraphics[width=\linewidth]{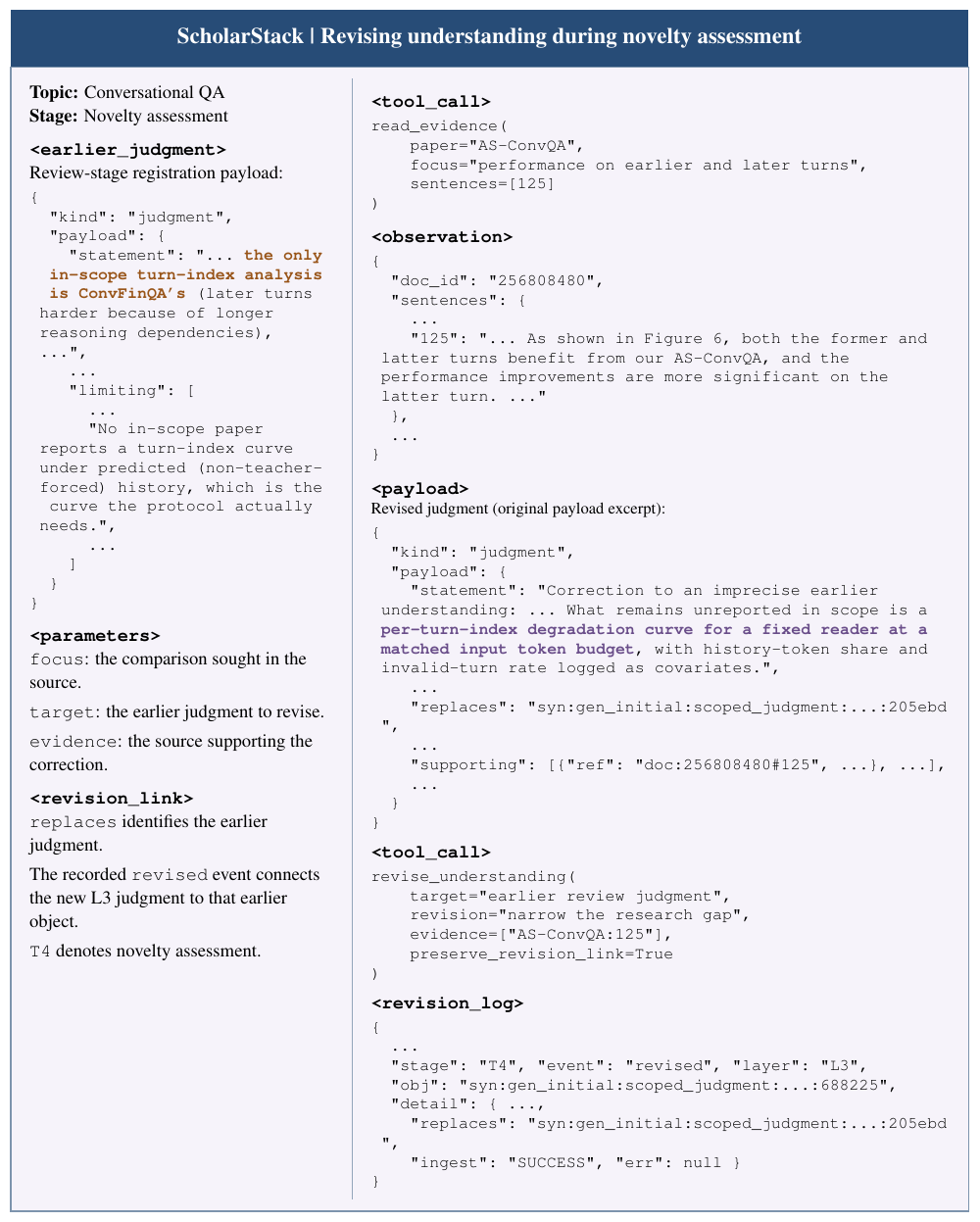}
\caption{An example of evidence-driven revision. The earlier review judgment appears on the left; the right shows counterevidence, the revised input payload, its write-back call, and the recorded L3 revision. Colored phrases highlight the corrected claim and refined scope. JSON records are excerpts from the actual run. Ellipses abbreviate records and identifiers.}
\label{fig:app-continuous-revision-trace}
\end{figure}
\clearpage
\endgroup

\end{document}